\documentclass[10pt,letterpaper]{article}
\usepackage[margin=1in]{geometry}
\usepackage{times}
\usepackage{microtype}
\usepackage[skip=3pt plus 1pt]{parskip}
\usepackage{amsmath,amssymb}
\usepackage[authoryear,round]{natbib}

\usepackage[T1]{fontenc}
\usepackage[utf8]{inputenc}
\usepackage{graphicx}
\usepackage{booktabs}
\usepackage[inline]{enumitem}
\usepackage{xcolor}
\usepackage{multirow}
\usepackage{pifont}
\usepackage{wrapfig}
\usepackage{float}
\usepackage{placeins}
\usepackage{caption}
\usepackage{fancyhdr}
\usepackage[hidelinks]{hyperref}
\usepackage[all]{pdfprivacy}
\floatstyle{ruled}
\newfloat{algorithm}{tbp}{loa}
\floatname{algorithm}{Algorithm}
\floatstyle{plain}

\definecolor{deltapos}{rgb}{0.18,0.48,0.28}
\definecolor{deltaneg}{rgb}{0.68,0.20,0.20}
\definecolor{deltaneu}{gray}{0.50}
\newcommand{\dpos}[1]{{\scriptsize\textcolor{deltapos}{(#1)}}}
\newcommand{\dneg}[1]{{\scriptsize\textcolor{deltaneg}{(#1)}}}
\newcommand{\dneu}[1]{{\scriptsize\textcolor{deltaneu}{(#1)}}}
\newcommand{\dchg}[2]{{\scriptsize\textcolor{#1}{(#2)}}}
\newcommand{\tablelayout}{\centering\setlength{\tabcolsep}{3pt}\renewcommand{\arraystretch}{1.06}}
\newcommand{\yes}{\ding{51}}
\newcommand{\no}{\ding{55}}

\title{\textbf{Learning from Recurrence:}\\[0.3em]
\Large Internalizing Depth-Induced Preferences in Looped Language Models}
\author{\textbf{Hui Chen}\\
  \small University of Wisconsin-Madison\\
  \small\href{mailto:hchen795@wisc.edu}{\nolinkurl{hchen795@wisc.edu}}
  \and
  \textbf{Joshua Elsdon}\\
  \small Microsoft\\
  \small\href{mailto:joshua@elsdon.io}{\nolinkurl{joshua@elsdon.io}}
  \and
  \textbf{Can Goksen}\\
  \small Microsoft\\
  \small\href{mailto:can.goksen@microsoft.com}{\nolinkurl{can.goksen@microsoft.com}}
}
\date{}

\begin{document}

\maketitle
\thispagestyle{fancy}

\begin{abstract}
A looped language model repeatedly applies the same block to its hidden state
before predicting each token. The number of passes is its \emph{recurrent depth}.
Increasing this depth gives the model more computation per token with the same
parameters. Can the shared block learn from the predictions it makes at greater
depth? Following prior work, we retrofit pretrained OLMo-2-1B with a shared
recurrent block and continue mid-training over varying recurrent
depths. In the four-pass model, additional passes mainly redistribute
probability among tokens that already rank highly after one pass. We use the
continuations generated at greater depth to train only the shared block through
a single pass, without reference answers or correctness filtering. We call
this transfer \emph{preference write-back}. Write-back further improves one-pass and multi-pass
accuracy on most tasks. At their target depths, the updated models outperform
a non-recurrent control matched for mid-training data and token budget on all
six mathematical reasoning benchmarks, with a mean relative gain in accuracy
of up to 46.2\% across tasks. At four-pass inference, one-pass training
outperforms full-depth training on the same targets in the model with input
injection. This advantage disappears in the matched model without injection.
On GSM8K, using the updated block only on passes two through four recovers
nearly the entire four-pass gain, showing that an update learned in one pass
can improve later applications of the same block.

\end{abstract}

\section{Introduction}

Looped Transformer language models scale inference-time computation by
reapplying recurrent blocks before emitting each token, decoupling computation
depth from layer depth
\citep{dehghani2019universal,saunshi2025reasoning,geiping2025scaling}.
This makes recurrent depth attractive both for serving---via parameter sharing,
depth-wise batching, and KV-cache reuse
\citep{lan2020albert,bae2025relaxed,bae2025mixture,geiping2025scaling}---and for latent test-time computation,
where the model iterates in hidden-state space before emitting a token
\citep{geiping2025scaling,mcleish2025teaching,
koishekenov2025encode,zhu2025scaling}.
Empirically, such models can improve with additional recurrent depth and, in
some settings, rival language models several times larger
\citep{saunshi2025reasoning,geiping2025scaling,zhu2025scaling}.

These gains, however, do not follow from weight sharing alone.
Although reapplying a shared block makes additional depth available,
the model must learn to use it across repeated passes.
Recent work has therefore developed recurrent architectures and
depth-training strategies toward this goal
\citep{geiping2025scaling,mcleish2025teaching,koishekenov2025encode}.
Building on these approaches, we retrofit a pretrained language model
with recurrence and adapt it through continued mid-training over
varying execution depths. Yet developing a model that benefits from
additional recurrence leaves a further question: what do these
additional passes actually change in its predictions?

Recurrent models can learn genuine iterative computation in controlled
settings, including repeated algorithmic operations and computing intermediate
results in one recurrent pass that are used in subsequent passes
\citep{schwarzschild2021can,bansal2022endtoend,saunshi2025reasoning,
jolicoeurmartineau2025trm,kohli2026loopthink,huang2026equilibrium}.
But these results do not determine the role of additional passes in pretrained
language models retrofitted with recurrence
\citep{bae2025relaxed,mcleish2025teaching,koishekenov2025encode}: improvements with depth may reflect
new intermediate computation, refinement of an already-formed predictive
state, or a combination of the two.

We therefore examine how the next-token distribution changes with recurrent
depth. Our intermediate readouts show that additional recurrence primarily
redistributes probability among tokens that already rank highly after
shallow execution.
The leading token often gains probability, while changes in the relative
ordering of nearby candidates can also alter which token the model favors.
These changes in preference can shape the continuation the model produces
and the context of subsequent predictions.
This observation connects to recent results on simple self-distillation,
where changes in a model's own predictive distribution can provide useful
training signal \citep{zhang2026ssd}.
Here, deeper recurrent execution itself produces these preference changes,
suggesting that the resulting behavior could guide further learning of the
shared recurrent operator without an additional correctness signal.

\textbf{This paper} tests this possibility by using trajectories generated
at deeper recurrent depth as supervision for the same model, while training
only the shared recurrent block through a shallower execution. We refer to
this process as \emph{preference write-back}. We take the shared recurrent
operator as the object of improvement, with its subsequent repeated use
as a central consideration. Shallow execution provides a path for learning
the update, while its value is assessed through subsequent applications
of the same operator.

Specifically, we make the following \textbf{contributions}. We show that
additional recurrence refines predictive preferences among plausible
continuations (Section~\ref{sec:depth-supervision}), and turn this
depth-induced behavior into supervision without reference answers or
filtering (Section~\ref{sec:method-writeback}). By separating teacher,
training and inference depth, we find that learning from deeper behavior
improves recurrent inference, with single-pass learning outperforming
full-depth learning on the same targets in our input-injected models
(Section~\ref{sec:training-depth}). We further show that the shallow-learned
update remains useful when the shared operator is applied at later
recurrent passes (Section~\ref{sec:core-position}). Controlled comparisons
reveal that the advantage of shallow learning depends on the recurrent
recipe, connecting how recurrent computation is developed to how it can
subsequently be improved (Section~\ref{sec:regime}). We also examine
the source of supervision and its transfer across question pools
(Section~\ref{sec:ablations}), together with the retention of capabilities
beyond mathematical reasoning (Section~\ref{sec:retention}).
\section{Learning from recurrence}
\label{sec:method}
\subsection{Looped language models and repeated computation across depth}
\label{sec:looped-lm}

\paragraph{A shared computation across depth.}
Looped language models increase execution depth by repeatedly applying the
same recurrent core to an evolving latent state. We develop this shared
computation through recurrent mid-training, then seek to improve the same core
using behavior revealed by deeper execution.

\paragraph{Input-conditioned recurrent computation.}
We retrofit the OLMo-2 1B stage-1 checkpoint \citep{olmo2} into a
non-recurrent input stack \(P\), a shared recurrent core \(F_\theta\), and a
non-recurrent output stack \(C\), following prior recurrent architectures
\citep{geiping2025scaling} and prior retrofits of pretrained models
\citep{mcleish2025teaching}.
The input and output stacks are termed the \emph{prelude} and \emph{coda}
in prior work. We assign the first seven Transformer blocks to \(P\), the
middle four to \(F_\theta\), and the final five to \(C\), following the
partition used for this model family \citep{koishekenov2025encode}.
This partition places recurrence in a middle region where the angular changes
between consecutive hidden representations have stabilized, retaining separate
input and output transformations.

Let \(x_{1:n}\in\mathcal{V}^{n}\) be a prefix of \(n\) tokens from vocabulary
\(\mathcal{V}\). The input stack, including token embeddings, produces
\(e=P(x_{1:n})\in\mathbb{R}^{n\times d}\), where \(d\) is the model width.
Starting from an initial state \(s_0\in\mathbb{R}^{n\times d}\), an execution
of depth \(r\) computes
\[
s_i =
F_\theta\!\left(A_\psi\!\left([s_{i-1};e]\right)\right),
\qquad i=1,\ldots,r .
\]
Here, \(s_i\in\mathbb{R}^{n\times d}\) is the state after pass \(i\), and
\([\cdot;\cdot]\) denotes concatenation along the hidden dimension.
The recurrent core
\(F_\theta:\mathbb{R}^{n\times d}\rightarrow\mathbb{R}^{n\times d}\)
and the position-wise linear projection
\(A_\psi:\mathbb{R}^{2d}\rightarrow\mathbb{R}^{d}\)
reuse the same parameters \(\theta\) and \(\psi\) at every pass.
This \emph{input injection} makes the same representation \(e\) available
throughout the evolution of the recurrent state.
After the final pass, the output stack
\(C:\mathbb{R}^{n\times d}\rightarrow\mathbb{R}^{n\times d}\)
and language-model head \(W\in\mathbb{R}^{|\mathcal{V}|\times d}\) produce
the next-token distribution
\(p_r(\cdot\mid x_{1:n})
=\operatorname{softmax}(W[C(s_r)]_n)\),
where \([\cdot]_n\in\mathbb{R}^{d}\) selects the final-position
representation. The input and output stacks are each applied once.

\paragraph{Learning across recurrent depths.}
We adapt the pretrained parameters to recurrent execution through continued
causal language-model training on the Dolmino mixture \citep{olmo2}.
During this stage, the execution depth is sampled once per training step
from a Poisson-lognormal depth distribution, following prior
recurrent-depth training \citep{geiping2025scaling}.
Let
\[
r \sim \pi_t,
\qquad
\mathbb{E}_{r\sim\pi_t}[r] = \mu_t ,
\]
where \(r\) is the sampled number of recurrent passes and \(\pi_t\) denotes the
Poisson-lognormal depth distribution at training step \(t\).
Rather than training immediately at the final depth distribution, we follow the
recurrence curriculum of \citet{mcleish2025teaching} and gradually increase
its mean \(\mu_t\) toward a target value \(m\).
The language-model objective is therefore optimized over recurrent computations
of varying depth, training the shared core to operate across the depth
distribution.

We call \(m\) the model's \emph{mean depth}, using target values of 4, 8,
and 32. Mean depth characterizes the recurrent-depth distribution under which
the model is learned, whereas the \emph{execution depth} \(r\) specifies the
number of recurrent passes used in an individual execution.
Once mid-training is complete, \(r\) remains independently selectable for a
fixed checkpoint, so shallow and deep executions are obtained by applying the
same learned recurrent computation for different numbers of passes.
Initialization and curriculum details are provided in
Appendix~\ref{sec:experimental-details}, and Appendix~\ref{sec:app-midtraining-progress} shows
that recurrence already improves accuracy during mid-training.
The resulting checkpoint supplies both the model to be updated and the deeper
executions from which we construct supervision.

\subsection{Recurrence-induced preferences as supervision}
\label{sec:depth-supervision}

\begin{wrapfigure}{r}{0.48\textwidth}
\centering
\includegraphics[width=\linewidth]{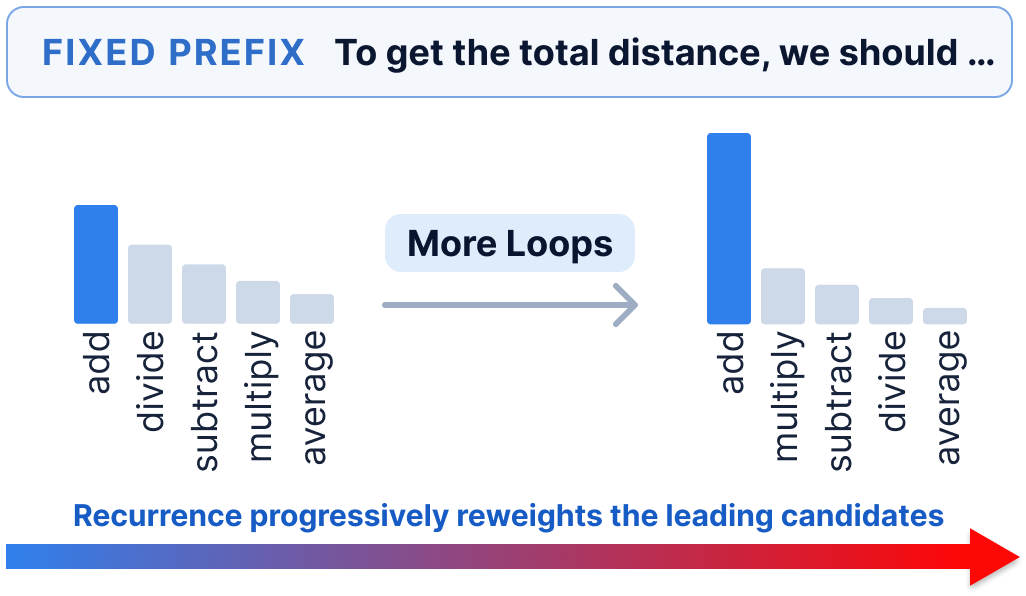}
\caption{\textbf{Recurrence revises predictive preference.}
At a fixed prefix, successive recurrent passes redistribute probability among leading candidates,
occasionally changing which continuation is preferred.}
\label{fig:motivation-reweighting}
\end{wrapfigure}

Once mid-training is complete, execution depth can be varied while the learned
parameters remain fixed.
We examine how additional recurrence changes the next-token distribution of a
fixed model, focusing on how repeated applications of the shared core reshape
its preferences among candidate continuations.
We first characterize this change, then use the preference of deeper execution
as supervision.

\paragraph{Recurrence refines predictive preference.}
At the level of intermediate predictions, additional recurrence primarily
refines preferences among already plausible continuations.
We expose this change using intermediate readouts along a fixed deep-generation
path, decoding each recurrent state with the same output stack while leaving the
generation itself unchanged.
In the mean-depth-4 model, the top-ranked token remains unchanged across four
passes at \(95.6\%\) of generated positions, even as probability shifts toward
the leading token while total top-10 mass changes little.
When the final top-ranked token does change, \(99.4\%\) of those choices are
already contained in the first-pass top five.
The dominant effect of recurrence is therefore a redistribution of probability
within the head of the predictive distribution, occasionally strong enough to
change the selected token.
These changes suggest using the choices made by deeper execution as
behavioral supervision.
Figure~\ref{fig:motivation-reweighting} illustrates this effect.
Appendix~\ref{sec:stable-decision-probability} details the readout protocol
and analysis.

\paragraph{From predictive preference to behavioral supervision.}
We next represent the preferences revealed by deeper recurrence as
learnable targets. A decoding rule turns the deeper next-token distribution
into a complete autoregressive continuation.
This preserves the token preferred at each step and the
contexts induced by those choices, through which subsequent
preferences are expressed.

In our default construction, we freeze the pre-update checkpoint and choose a
teacher depth \(r_{\mathrm T}\).
For a prompt \(x\), greedy decoding produces a continuation
\(\hat y=(\hat y_1,\ldots,\hat y_T)\) according to
\[
\hat y_t
=
\arg\max_{v\in\mathcal V}
p^{(0)}_{r_{\mathrm T}}
\!\left(v\mid x,\hat y_{<t}\right),
\qquad t=1,\ldots,T,
\]
where \(p^{(0)}_{r_{\mathrm T}}\) is the predictive distribution of the frozen
pre-update model after \(r_{\mathrm T}\) recurrent passes.
The resulting trajectory is a sequence-level realization of the preferences
expressed by deeper execution. It retains the selected tokens and the contexts
they create, without preserving the full predictive distribution.

For \(N\) training prompts, we collect one such continuation per prompt to form
\[
\mathcal D_{r_{\mathrm T}}
=
\left\{
\left(x^{(j)},\hat y^{(j)}\right)
\right\}_{j=1}^{N}.
\]
Each target retains the complete generated continuation, including its
reasoning and final answer.
Reference answers to the target questions are neither provided during
generation nor used to filter trajectories.
Teacher depth determines which recurrent behavior is exposed, while the
trajectory-construction rule determines which realization of that behavior is
retained as supervision.
How this supervision is written back into the shared computation is
a separate choice, discussed next.

\subsection{Writing preferences back into the shared recurrent core}
\label{sec:method-writeback}

Once deeper behavior has been converted into supervision, the remaining
problem is to turn a transient effect of additional computation into a
persistent change in the computation itself.
In a weight-tied model, this means writing the behavior back into the shared
operator that produced it and that subsequent recurrence will reuse.
Crucially, the execution that reveals the behavior need not be the execution
through which the operator learns it.
We refer to this transfer from depth-induced behavior into reusable computation
as \emph{preference write-back} (Figure~\ref{fig:framework} and
Algorithm~\ref{alg:writeback} in Appendix~\ref{sec:app-self-distillation}).

\begin{figure}[tb]
\centering
\includegraphics[width=\linewidth]{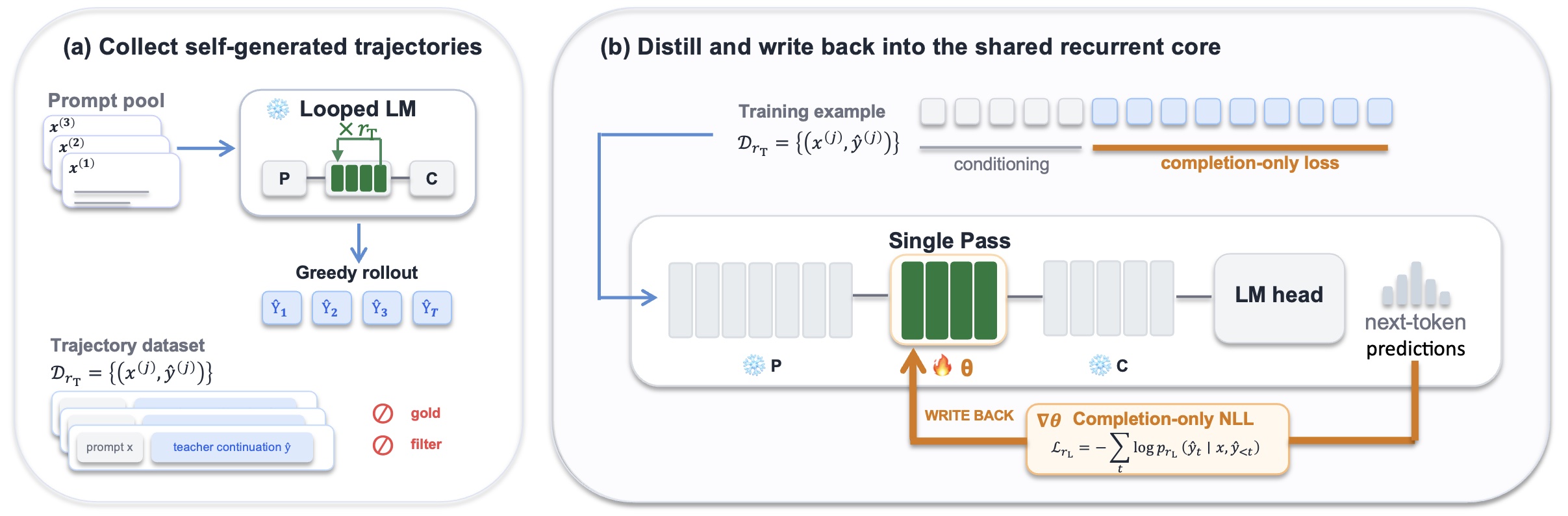}
\caption{\textbf{Preference write-back.} (a) The frozen model generates one greedy continuation per
prompt at teacher depth \(r_{\mathrm T}\), without reference answers or filtering. (b) These
trajectories train only the shared recurrent core through a single pass
(\(r_{\mathrm L}{=}1\)). The updated core is reused at every pass.}
\label{fig:framework}
\end{figure}

\paragraph{Decoupling behavior exposure from incorporation.}
Preference write-back separates the source of supervision from the computation
through which that supervision is learned.
The \emph{teacher depth} \(r_{\mathrm T}\) specifies the execution whose
behavior is retained as supervision, while the \emph{training depth}
\(r_{\mathrm L}\) specifies the recurrent computation through which that
supervision acts on the shared operator.
For a fixed target set \(\mathcal{D}_{r_{\mathrm T}}\), changing
\(r_{\mathrm L}\) leaves both the behavioral target and the shared
parameterization unchanged. Only the computational path connecting that
target to the shared parameters changes.
Behavior exposed at one depth can thus be learned through another
execution of the same operator.

\paragraph{Writing behavior into the reusable operator.}
The supervision is behavioral, but its write-back destination is the shared
recurrent operator.
Let \(\theta_0\) denote the pre-update parameters of the recurrent core, and
let \(\phi\) collect the remaining model parameters, which are held fixed.
\(\phi\) includes the injection projection \(A_{\psi}\), which acts in the
forward computation but is not updated.
Given \(\mathcal{D}_{r_{\mathrm T}}\), we initialize
\(\theta=\theta_0\) and optimize the recurrent core through an execution of
depth \(r_{\mathrm L}\) using completion-only next-token likelihood.
\[
\mathcal{L}_{r_{\mathrm L}}(\theta)
=
-\frac{1}{\sum_j T_j}
\sum_{j=1}^{N}
\sum_{t=1}^{T_j}
\log
p_{r_{\mathrm L}}
\!\left(
\hat y_t^{(j)}
\mid
x^{(j)},\hat y_{<t}^{(j)};
\theta,\phi
\right).
\]
Here,
\(p_{r_{\mathrm L}}(\cdot\,;\theta,\phi)\) denotes the predictive distribution
obtained after \(r_{\mathrm L}\) applications of the recurrent core, in the
notation of Section~\ref{sec:looped-lm}.
Only the shared recurrent core is updated. All remaining model parameters
stay fixed.
The resulting parameters \(\theta_1\) store the deep-execution behavior in the
same transformation that recurrence will invoke again.
Training depth controls the path by which supervision reaches the
operator, while the object being changed remains the reusable computation.

\paragraph{Reusing the written-back computation.}
After optimization, the updated core \(F_{\theta_1}\) becomes the recurrent
operator used at every subsequent pass, while the remaining parameters
\(\phi\) remain unchanged.
At inference depth \(r_{\mathrm E}\), the model repeatedly applies
\(F_{\theta_1}\) as the recurrent state evolves.
This reuse follows from weight sharing. Whether an update learned through
shallow execution remains useful on states reached later in recurrence is a
separate empirical question.

Preference write-back thus separates the depth that exposes behavior from
the depth used to learn it and the depth at which the updated computation is
reused.
These roles are indexed by teacher depth \(r_{\mathrm T}\), training depth
\(r_{\mathrm L}\), and inference depth \(r_{\mathrm E}\), respectively, while
all three operate on the same shared recurrent computation.
We therefore assess the update through its repeated use at inference,
not only through the execution used to learn it.
\section{Preference write-back in looped language models}
\label{sec:experiments}

\paragraph{Experimental setup.}
We study looped language models obtained by continuing mid-training an OLMo-2-1B base
model toward mean recurrent depths of 4, 8, and 32. Unless otherwise stated,
self-distillation follows the recipe introduced in Section~\ref{sec:method}.
Each model generates greedy trajectories at its standard deeper execution depth,
without correctness filtering, and only the shared recurrent block is updated
through one recurrent pass. We compare the looped checkpoints with a non-looped
control trained on the same data and 50B mid-training tokens. We evaluate each
looped model before and after write-back at one pass and at its standard inference
depth, focusing on mathematical reasoning under greedy decoding. Architecture,
training, and evaluation details are given in Section~\ref{sec:method} and
Appendix~\ref{sec:experimental-details}.

\subsection{Learning from deeper recurrence improves recurrent inference}
\label{sec:writeback}

\textbf{The looped checkpoints already benefit from additional recurrence
before write-back.} At their mean depths, all three outperform the non-looped
control on each of the six mathematical reasoning benchmarks, with mean
per-task relative gains of 26.5--39.9\% (Table~\ref{tab:main}). On GSM8K,
the mean-depth-4 checkpoint improves from 40.86\% after one pass to 56.18\%
after four passes.

Learning from its own four-pass continuations raises one-pass accuracy to
48.52\% and four-pass accuracy to 62.70\%. The updated model continues to
benefit from additional recurrence, gaining 14.18 points from one to four
passes. Across the three models, write-back increases accuracy in 15 of the 18
model--benchmark comparisons at their mean depths, including all six for the
mean-depth-4 model. GSM8K, GSM1K, GSM-Symbolic, and ASDiv improve at all
three mean depths.
The final models achieve mean per-task relative gains of 34.9--46.2\% over
the non-looped control.

\textbf{Additional recurrence supplies supervision beyond one-pass
self-training.} On the mean-depth-4 model, we use the same question pool,
one-pass learning, and four-pass inference, varying only the teacher depth
(Table~\ref{tab:ablations}). Four-pass teacher continuations yield 62.70\%
on GSM8K and 54.36\% on GSM1K, compared with 59.29\% and 51.95\% from
one-pass teacher continuations.

\begin{table}[tb]
\caption{\textbf{Mathematical reasoning before and after preference write-back across recurrent depths.}
Exact match (\%) under greedy decoding at depth 1 and at the mean depth \(m\).
\emph{Mean depth} rows report the looped checkpoints before write-back. The write-back
rows update only the shared recurrent block using greedy depth-\(m\) trajectories
(50,000 OpenMath questions for \(m{=}4\), 10,000 otherwise). Parenthesized values
and the last column report relative changes at depth \(m\) against the non-looped
control.}
\label{tab:main}
\tablelayout
\resizebox{\textwidth}{!}{\begin{tabular}{@{}lrrrrrrrrrrrrc@{}}
\toprule
Model & \multicolumn{2}{c}{GSM8K} & \multicolumn{2}{c}{GSM1K} & \multicolumn{2}{c}{SVAMP} & \multicolumn{2}{c}{GSM-Symbolic} & \multicolumn{2}{c}{ASDiv} & \multicolumn{2}{c}{AQuA-RAT} & Avg. \\
\cmidrule(lr){2-3} \cmidrule(lr){4-5} \cmidrule(lr){6-7} \cmidrule(lr){8-9} \cmidrule(lr){10-11} \cmidrule(lr){12-13}
\multicolumn{1}{r}{Depth} & \multicolumn{1}{c}{1} & \multicolumn{1}{c}{\(m\)} & \multicolumn{1}{c}{1} & \multicolumn{1}{c}{\(m\)} & \multicolumn{1}{c}{1} & \multicolumn{1}{c}{\(m\)} & \multicolumn{1}{c}{1} & \multicolumn{1}{c}{\(m\)} & \multicolumn{1}{c}{1} & \multicolumn{1}{c}{\(m\)} & \multicolumn{1}{c}{1} & \multicolumn{1}{c}{\(m\)} & vs.\ control \\
\midrule
Non-looped control & 44.05 & -- & 39.83 & -- & 56.50 & -- & 37.42 & -- & 59.39 & -- & 10.24 & -- & -- \\
\midrule
Mean depth 4 & 40.86 & 56.18 & 36.93 & 51.12 & 52.40 & 63.30 & 31.82 & 43.56 & 55.44 & 64.82 & 14.17 & 16.93 & \textcolor{deltapos}{+26.5\%} \\
 &  & \dpos{+27.5\%} &  & \dpos{+28.3\%} &  & \dpos{+12.0\%} &  & \dpos{+16.4\%} &  & \dpos{+9.1\%} &  & \dpos{+65.4\%} &  \\
+ Self-distillation & 48.52 & 62.70 & 44.65 & 54.36 & 56.40 & 64.70 & 36.92 & 47.64 & 59.09 & 66.55 & 12.20 & 18.11 & \textcolor{deltapos}{+34.9\%} \\
 &  & \dpos{+42.3\%} &  & \dpos{+36.5\%} &  & \dpos{+14.5\%} &  & \dpos{+27.3\%} &  & \dpos{+12.1\%} &  & \dpos{+76.9\%} &  \\
\midrule
Mean depth 8 & 22.59 & 56.33 & 20.50 & 52.53 & 39.90 & 66.00 & 15.76 & 45.06 & 46.33 & 66.20 & 10.24 & 23.62 & \textcolor{deltapos}{+39.9\%} \\
 &  & \dpos{+27.9\%} &  & \dpos{+31.9\%} &  & \dpos{+16.8\%} &  & \dpos{+20.4\%} &  & \dpos{+11.5\%} &  & \dpos{+130.8\%} &  \\
+ Self-distillation & 37.38 & 62.85 & 32.37 & 56.02 & 43.60 & 65.20 & 27.94 & 48.42 & 51.84 & 67.90 & 5.91 & 22.44 & \textcolor{deltapos}{+43.6\%} \\
 &  & \dpos{+42.7\%} &  & \dpos{+40.6\%} &  & \dpos{+15.4\%} &  & \dpos{+29.4\%} &  & \dpos{+14.3\%} &  & \dpos{+119.2\%} &  \\
\midrule
Mean depth 32 & 4.09 & 57.01 & 3.24 & 53.36 & 19.70 & 67.70 & 2.68 & 44.90 & 20.39 & 66.55 & 0.79 & 19.29 & \textcolor{deltapos}{+34.0\%} \\
 &  & \dpos{+29.4\%} &  & \dpos{+34.0\%} &  & \dpos{+19.8\%} &  & \dpos{+20.0\%} &  & \dpos{+12.1\%} &  & \dpos{+88.5\%} &  \\
+ Self-distillation & 31.84 & 62.09 & 26.39 & 54.36 & 36.20 & 65.70 & 20.66 & 48.08 & 44.95 & 67.07 & 2.76 & 24.80 & \textcolor{deltapos}{+46.2\%} \\
 &  & \dpos{+41.0\%} &  & \dpos{+36.5\%} &  & \dpos{+16.3\%} &  & \dpos{+28.5\%} &  & \dpos{+12.9\%} &  & \dpos{+142.3\%} &  \\
\bottomrule
\end{tabular}
}
\end{table}

\textbf{Write-back further improves the looped checkpoint under a common
learning budget.} We compare gold-answer SFT, SSD-style self-training
\citep{zhang2026ssd}, and distillation on a common random subset of 1,024
GSM8K training questions
(Table~\ref{tab:budget-comparison}). Both starting checkpoints share the
pretrained initialization, mid-training data, data order, and 50B-token budget.
The primary comparisons use a common cap of approximately
\(1.575\times 10^{16}\) estimated matrix-multiplication FLOPs for trajectory
generation, teacher scoring, and student training. Sequence KD
\citep{kim2016sequence} learns from the same four-pass continuations as
write-back and gives the best non-looped scores of 47.46\% on GSM8K and
40.83\% on GSM1K. The looped checkpoint already reaches 56.18\% and
51.12\% at four-pass inference.

On the looped checkpoint, DistiLLM-2 \citep{ko2025distillm2}, GKD
\citep{agarwal2024gkd}, and EOPD \citep{jin2026eopd} use the same original
four-pass teacher as write-back. Write-back achieves the highest four-pass
accuracy on both tasks among the tested updates, at 60.50\% and 51.54\%.
Its GSM8K gain of 4.32 points is more than twice the 2.12-point gain of
DistiLLM-2, the strongest tested KD alternative on this task.

\begin{table}[!tbp]
\caption{\textbf{Write-back under a common learning budget.}
Exact match (\%) at inference depths 1 and 4 after learning on the same
1,024 GSM8K training questions.
\(B_{50}\) and \(L_0\) denote the non-looped and mean-depth-4 checkpoints.
\(L_0\) paths and teacher refer to its four-pass continuations and predictive
supervision, respectively. Gold denotes human solutions. Bold marks the
highest score within each checkpoint block at each depth.}
\label{tab:budget-comparison}
\tablelayout
\small
\begin{tabular*}{\linewidth}{@{\extracolsep{\fill}}llccrrrr@{}}
\toprule
Model & Supervision & Trainable & Learning & \multicolumn{2}{c}{GSM8K} & \multicolumn{2}{c}{GSM1K} \\
\cmidrule(lr){5-6}\cmidrule(l){7-8}
 & & parameters & depth & 1 & 4 & 1 & 4 \\
\midrule
Non-looped ($B_{50}$) & -- & -- & -- & 44.05 & -- & 39.83 & -- \\
+ SSD self-training & Own samples & Full & 1 & 42.23 & -- & 37.43 & -- \\
+ Gold-answer SFT & Gold & Full & 1 & 44.73 & -- & 40.17 & -- \\
+ Sequence KD & $L_0$ paths & Full & 1 & \textbf{47.46} & -- & \textbf{40.83} & -- \\
+ EOPD & $L_0$ teacher & Full & 1 & 45.72 & -- & 39.50 & -- \\
\midrule
Mean depth 4 ($L_0$) & -- & -- & -- & 40.86 & 56.18 & 36.93 & 51.12 \\
+ Gold-answer SFT & Gold & Core & 1 & 38.89 & 55.04 & 33.03 & 50.79 \\
 & Gold & Core & 4 & 38.51 & 56.10 & 34.77 & 51.20 \\
 & Gold & Full & 1 & 38.36 & 54.28 & 35.10 & 46.89 \\
 & Gold & Full & 4 & 40.03 & 55.50 & 33.03 & 47.72 \\
+ DistiLLM-2 & $L_0$ teacher & Core & 1 & 42.46 & 58.30 & 36.93 & 50.71 \\
+ GKD (on-policy) & $L_0$ teacher & Core & 1 & 41.93 & 56.71 & 36.76 & 49.05 \\
+ EOPD & $L_0$ teacher & Core & 1 & 41.70 & 56.86 & 36.51 & 50.46 \\
 & $L_0$ teacher & Full & 1 & 40.71 & 55.80 & 34.11 & 46.56 \\
+ Preference write-back & $L_0$ paths & Core & 1 & \textbf{42.76} & \textbf{60.50} & \textbf{39.50} & \textbf{51.54} \\
\bottomrule
\end{tabular*}

\end{table}

\subsection{Learning depth shapes the update to the shared operator}
\label{sec:training-depth}

\textbf{The depth used for learning matters even when the teacher
trajectories are fixed.} We hold the targets, objective, trainable parameters,
and optimization settings fixed, changing only the number of recurrent passes
used during learning (Table~\ref{tab:mechanism}). On the mean-depth-4 model,
learning four-pass targets through one pass reaches 62.70\% GSM8K accuracy
at four-pass inference, compared with 56.86\% after four-pass learning. The
corresponding advantages are 3.32 points on GSM1K and 2.50 points on SVAMP.
The same ordering holds on GSM8K and GSM1K at the mean depths of 8 and 32.
The mean-depth-32 checkpoint scores only 4.09\% on GSM8K after one pass,
yet learning through that pass raises its 32-pass accuracy from 57.01\%
to 62.09\%.

Training through only the final pass of a four-pass execution reaches
56.63\% on GSM8K in the mean-depth-4 model. Both this setting and one-pass
learning differentiate a single application of the core. The forward depth
used during learning therefore matters even when backpropagation covers
only one recurrent pass.

\begin{figure}[tb]
\noindent
\begin{minipage}[t]{0.56\linewidth}
\captionsetup{position=top,skip=0pt}
\caption{\textbf{Recurrent depth weakens and redirects the gradient reaching the first-pass state.}
(a) Cosine similarity to the one-pass gradient and relative norm at the first-pass state, with
the loss placed after one to four passes.
(b) Each pass's share of the sum of per-pass parameter-gradient norms under four-pass training.
Setup in Appendix~\ref{sec:gradient-diagnostic}.}
\label{fig:gradient-geometry}
\end{minipage}\hfill
\begin{minipage}[t]{0.41\linewidth}
\captionsetup{type=table,position=top,skip=0pt}
\caption{\textbf{Applying the shallow-learned block at later passes recovers nearly all of the four-pass gain.}
GSM8K exact match (\%) at four-pass inference when each pass uses the original (\no) or
the self-distilled (\yes) shared block. Intervals and per-question counts are in
Table~\ref{tab:app-core-replacement}.}
\label{tab:core-position}
\end{minipage}
\par\vspace{5pt}
\noindent
\begin{minipage}[b]{0.56\linewidth}
\centering
\includegraphics[height=1.4in]{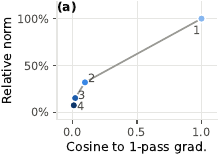}\hspace{0.5em}\includegraphics[height=1.4in]{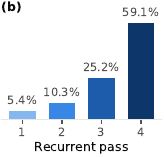}
\end{minipage}\hfill
\begin{minipage}[b]{0.41\linewidth}
\tablelayout
\small
\renewcommand{\arraystretch}{1.24}
\raisebox{\dimexpr\depth+1.5pt\relax}{\begin{tabular*}{\linewidth}{@{\extracolsep{\fill}}ccccr@{\extracolsep{0pt}\,}l@{}}
\toprule
\multicolumn{4}{c}{Pass} & & \\
\cmidrule(lr){1-4}
1 & 2 & 3 & 4 & \multicolumn{2}{c@{}}{Accuracy (\%)} \\
\midrule
\no & \no & \no & \no & 56.18 & \\
\yes & \no & \no & \no & 56.25 & \dneu{+0.1\%} \\
\no & \yes & \yes & \yes & 62.47 & \dpos{+11.2\%} \\
\yes & \yes & \yes & \yes & 62.70 & \dpos{+11.6\%} \\
\bottomrule
\end{tabular*}
\unskip}
\end{minipage}
\end{figure}

One-pass learning places the loss directly on the first-pass state. When
the loss is read out after four passes, the gradient at that same state
has 7.3\% of the norm of the direct one-pass gradient and a cosine similarity
of 0.009 (Figure~\ref{fig:gradient-geometry}a). At the parameter level, the
first and fourth applications account for 5.4\% and 59.1\% of the sum of
per-pass gradient norms under four-pass learning, respectively
(Figure~\ref{fig:gradient-geometry}b). Appendix~\ref{sec:app-depth-profile}
extends the performance comparison across inference depths and to a
mean-depth-2 model.

\begin{table}[tb]
\caption{\textbf{Single-pass learning outperforms full-depth learning on GSM8K and GSM1K at the three mean depths.}
Exact match (\%) at depth 1 and at the mean depth \(m\). Each block trains on the model's own
greedy depth-\(m\) trajectories (\emph{Teacher}) and varies the depth used during learning
(\emph{Training}). The \(4^\dagger\) setting uses four passes with gradients through the
last pass only. Changes are relative to the block's \emph{Mean depth} row.}
\label{tab:mechanism}
\tablelayout
\resizebox{\textwidth}{!}{\begin{tabular}{@{}lccllllllc@{}}
\toprule
Model & \multicolumn{2}{c}{Depth} & \multicolumn{2}{c}{GSM8K} & \multicolumn{2}{c}{GSM1K} & \multicolumn{2}{c}{SVAMP} & Avg. \\
\cmidrule(lr){2-3} \cmidrule(lr){4-5} \cmidrule(lr){6-7} \cmidrule(lr){8-9}
& Teacher & Training & Depth 1 & Depth \(m\) & Depth 1 & Depth \(m\) & Depth 1 & Depth \(m\) & gain \\
\midrule
Mean depth 4 & -- & -- & 40.86 & 56.18 & 36.93 & 51.12 & 52.40 & 63.30 & -- \\
\multirow{3}{*}{+ Self-distillation} & 4 & 1 & 48.52 & \textbf{62.70}\,\dpos{+11.6\%} & 44.65 & \textbf{54.36}\,\dpos{+6.3\%} & 56.40 & \textbf{64.70}\,\dpos{+2.2\%} & \textcolor{deltapos}{+6.7\%} \\
 & 4 & 4 & 37.83 & 56.86\,\dpos{+1.2\%} & 37.18 & 51.04\,\dneg{$-$0.2\%} & 52.00 & 62.20\,\dneg{$-$1.7\%} & \textcolor{deltaneg}{$-$0.2\%} \\
 & 4 & 4$^\dagger$ & 37.53 & 56.63\,\dpos{+0.8\%} & 35.44 & 49.96\,\dneg{$-$2.3\%} & 51.50 & 61.30\,\dneg{$-$3.2\%} & \textcolor{deltaneg}{$-$1.5\%} \\
\addlinespace[2pt]
Mean depth 8 & -- & -- & 22.59 & 56.33 & 20.50 & 52.53 & 39.90 & \textbf{66.00} & -- \\
\multirow{2}{*}{+ Self-distillation} & 8 & 1 & 37.38 & \textbf{62.85}\,\dpos{+11.6\%} & 32.37 & \textbf{56.02}\,\dpos{+6.6\%} & 43.60 & 65.20\,\dneg{$-$1.2\%} & \textcolor{deltapos}{+5.7\%} \\
 & 8 & 8 & 22.59 & 58.68\,\dpos{+4.2\%} & 19.67 & 51.87\,\dneg{$-$1.3\%} & 39.90 & 65.70\,\dneg{$-$0.5\%} & \textcolor{deltapos}{+0.8\%} \\
\addlinespace[2pt]
Mean depth 32 & -- & -- & 4.09 & 57.01 & 3.24 & 53.36 & 19.70 & \textbf{67.70} & -- \\
\multirow{2}{*}{+ Self-distillation} & 32 & 1 & 31.84 & \textbf{62.09}\,\dpos{+8.9\%} & 26.39 & \textbf{54.36}\,\dpos{+1.9\%} & 36.20 & 65.70\,\dneg{$-$3.0\%} & \textcolor{deltapos}{+2.6\%} \\
 & 32 & 32 & 4.93 & 58.76\,\dpos{+3.1\%} & 3.57 & 53.28\,\dneg{$-$0.2\%} & 19.80 & 65.30\,\dneg{$-$3.5\%} & \textcolor{deltaneg}{$-$0.2\%} \\
\bottomrule
\end{tabular}
}
\end{table}

\subsection{Shallow-learned updates remain useful in later recurrence}
\label{sec:core-position}

\textbf{Does an update learned through one pass improve later applications
of the same shared operator?} In the mean-depth-4 model, we choose at each
of four recurrent passes whether to use the original or self-distilled
shared block, while keeping the rest of the model fixed
(Table~\ref{tab:core-position}). Using the updated block only on the first
pass gives 56.25\% GSM8K accuracy, close to the original model's 56.18\%.
Keeping the original block on pass 1 and using the updated block on
passes 2--4 reaches 62.47\%, compared with 62.70\% when the update is
used throughout. The later-pass replacement recovers 6.29 of the 6.52-point
gain of the fully updated model. The update learned through one pass therefore
remains useful on states reached later in recurrence.

\subsection{Recurrent recipes shape subsequent learning}
\label{sec:regime}

\textbf{The advantage of shallow learning depends on the recurrent recipe.}
We compare mean-depth-4 models with and without input injection, starting
from the same pretrained checkpoint and following matched mid-training
procedures (Table~\ref{tab:regime}). In the shared-target comparison,
both models learn from the injected model's four-pass continuations.

With input injection, one-pass learning outperforms four-pass learning by
5.84 points on GSM8K and 3.32 points on GSM1K at four-pass inference.
Without injection, four-pass learning gives comparable GSM8K accuracy
(54.81\% versus 54.74\%) and higher accuracy on GSM1K (47.63\% versus
46.39\%) and SVAMP (59.60\% versus 56.80\%). How recurrent computation
is developed therefore matters for how it can subsequently be improved.

\begin{table}[tb]
\caption{\textbf{The advantage of shallow training depends on the recurrent recipe.}
Exact match (\%) for matched mean-depth-4 models with and without input injection, self-distilled
through one or four passes on their own greedy four-pass trajectories (\emph{own}) or on those of
the injected model (\emph{injected}). Changes are relative to the \emph{Mean depth 4} row with the
same injection setting.}
\label{tab:regime}
\tablelayout
\resizebox{\textwidth}{!}{\begin{tabular}{@{}lcccllllllc@{}}
\toprule
Model & Injection & Trajectories & Training depth & \multicolumn{2}{c}{GSM8K} & \multicolumn{2}{c}{GSM1K} & \multicolumn{2}{c}{SVAMP} & Avg. \\
\cmidrule(lr){5-6} \cmidrule(lr){7-8} \cmidrule(lr){9-10}
& & &  & Depth 1 & Depth \(m\) & Depth 1 & Depth \(m\) & Depth 1 & Depth \(m\) & gain \\
\midrule
Mean depth 4 & \yes & -- & -- & 40.86 & 56.18 & 36.93 & 51.12 & 52.40 & 63.30 & -- \\
\multirow{2}{*}{+ Self-distillation} & \yes & own & 1 & 48.52 & \textbf{62.70}\,\dpos{+11.6\%} & 44.65 & \textbf{54.36}\,\dpos{+6.3\%} & 56.40 & \textbf{64.70}\,\dpos{+2.2\%} & \textcolor{deltapos}{+6.7\%} \\
 & \yes & own & 4 & 37.83 & 56.86\,\dpos{+1.2\%} & 37.18 & 51.04\,\dneg{$-$0.2\%} & 52.00 & 62.20\,\dneg{$-$1.7\%} & \textcolor{deltaneg}{$-$0.2\%} \\
\midrule
Mean depth 4 & \no & -- & -- & 43.90 & 50.80 & 38.42 & 45.31 & 54.90 & 58.70 & -- \\
\multirow{4}{*}{+ Self-distillation} & \no & own & 1 & 47.16 & 52.39\,\dpos{+3.1\%} & 42.90 & 45.31\,\dneu{+0.0\%} & 54.60 & 56.50\,\dneg{$-$3.7\%} & \textcolor{deltaneg}{$-$0.2\%} \\
 & \no & own & 4 & 44.58 & 50.72\,\dneg{$-$0.1\%} & 40.41 & 46.72\,\dpos{+3.1\%} & 53.90 & 59.00\,\dpos{+0.5\%} & \textcolor{deltapos}{+1.2\%} \\
 & \no & injected & 1 & 48.98 & 54.74\,\dpos{+7.8\%} & 41.91 & 46.39\,\dpos{+2.4\%} & 56.40 & 56.80\,\dneg{$-$3.2\%} & \textcolor{deltapos}{+2.3\%} \\
 & \no & injected & 4 & 45.26 & 54.81\,\dpos{+7.9\%} & 39.75 & 47.63\,\dpos{+5.1\%} & 56.20 & 59.60\,\dpos{+1.5\%} & \textcolor{deltapos}{+4.9\%} \\
\bottomrule
\end{tabular}
}
\end{table}

\begin{table}[!tbp]
\caption{\textbf{One-pass write-back across recurrent model families.}
Exact match (\%) at depth 1 and the reported deeper depth \(m\). Preference
write-back uses 1,024 GSM8K training questions. Parentheses show relative changes
from the original checkpoint at the same depth, computed from correct-answer counts.
N/A marks a zero baseline. Nonzero baselines below 0.1\% represent one correct answer.
Bold marks the higher score in each comparison.}
\label{tab:public-models}
\tablelayout
\small
\newcommand{\publicmodelscore}[2]{#1\,\makebox[4.35em][r]{#2}}
\begin{tabular*}{\linewidth}{@{\extracolsep{\fill}}lclrrrr@{}}
\toprule
Model & \(m\) & Checkpoint & \multicolumn{2}{c}{GSM8K} & \multicolumn{2}{c}{GSM1K} \\
\cmidrule(lr){4-5}\cmidrule(l){6-7}
 & & & \multicolumn{1}{c}{Depth 1} & \multicolumn{1}{c}{Depth \(m\)} & \multicolumn{1}{c}{Depth 1} & \multicolumn{1}{c}{Depth \(m\)} \\
\midrule
Mean depth 4 (ours) & 4 & Original & \publicmodelscore{40.86}{} & \publicmodelscore{56.18}{} & \publicmodelscore{36.93}{} & \publicmodelscore{51.12}{} \\
 &  & Write-back & \publicmodelscore{\textbf{42.76}}{\dpos{+4.6\%}} & \publicmodelscore{\textbf{60.50}}{\dpos{+7.7\%}} & \publicmodelscore{\textbf{39.50}}{\dpos{+7.0\%}} & \publicmodelscore{\textbf{51.54}}{\dpos{+0.8\%}} \\
\addlinespace[3pt]
Retro-OLMo & 16 & Original & \publicmodelscore{29.49}{} & \publicmodelscore{45.87}{} & \publicmodelscore{24.65}{} & \publicmodelscore{43.07}{} \\
 &  & Write-back & \publicmodelscore{\textbf{33.97}}{\dpos{+15.2\%}} & \publicmodelscore{\textbf{48.60}}{\dpos{+6.0\%}} & \publicmodelscore{\textbf{30.04}}{\dpos{+21.9\%}} & \publicmodelscore{\textbf{45.23}}{\dpos{+5.0\%}} \\
\addlinespace[3pt]
Retro-LLaMA & 8 & Original & \publicmodelscore{25.63}{} & \publicmodelscore{46.17}{} & \publicmodelscore{24.15}{} & \publicmodelscore{43.98}{} \\
 &  & Write-back & \publicmodelscore{\textbf{32.22}}{\dpos{+25.7\%}} & \publicmodelscore{\textbf{48.60}}{\dpos{+5.3\%}} & \publicmodelscore{\textbf{31.29}}{\dpos{+29.6\%}} & \publicmodelscore{\textbf{44.32}}{\dpos{+0.8\%}} \\
\addlinespace[3pt]
Huginn (3.5B) & 32 & Original & \publicmodelscore{0.00}{} & \publicmodelscore{32.37}{} & \publicmodelscore{0.08}{} & \publicmodelscore{28.46}{} \\
 &  & Write-back & \publicmodelscore{\textbf{1.59}}{\dneu{N/A}} & \publicmodelscore{\textbf{32.68}}{\dpos{+0.9\%}} & \publicmodelscore{\textbf{1.58}}{\dpos{+1,800.0\%}} & \publicmodelscore{\textbf{28.80}}{\dpos{+1.2\%}} \\
\addlinespace[3pt]
LoopUS & 8 & Original & \publicmodelscore{0.08}{} & \publicmodelscore{\textbf{48.82}}{} & \publicmodelscore{0.08}{} & \publicmodelscore{48.22}{} \\
 &  & Write-back & \publicmodelscore{\textbf{55.65}}{\dpos{+73,300.0\%}} & \publicmodelscore{48.67}{\dneg{$-$0.3\%}} & \publicmodelscore{\textbf{53.53}}{\dpos{+64,400.0\%}} & \publicmodelscore{\textbf{48.46}}{\dpos{+0.5\%}} \\
\addlinespace[3pt]
Ouro & 4 & Original & \publicmodelscore{26.16}{} & \publicmodelscore{\textbf{79.68}}{} & \publicmodelscore{25.06}{} & \publicmodelscore{\textbf{77.43}}{} \\
 &  & Write-back & \publicmodelscore{\textbf{38.67}}{\dpos{+47.8\%}} & \publicmodelscore{77.18}{\dneg{$-$3.1\%}} & \publicmodelscore{\textbf{36.10}}{\dpos{+44.0\%}} & \publicmodelscore{74.69}{\dneg{$-$3.5\%}} \\
\bottomrule
\end{tabular*}

\end{table}

\textbf{Shallow-learned updates also benefit deeper inference in other
recurrent models.} We apply write-back to five public looped checkpoints
using the same 1,024 GSM8K training questions. Each model learns from its own
deeper continuations by updating the shared core through one pass
(Table~\ref{tab:public-models}). Retro-OLMo and Retro-LLaMA
\citep{mcleish2025teaching} gain 2.73 and 2.43 points on GSM8K at 16 and 8
passes, respectively, and improve on GSM1K. Together with our
model and the 3.5B Huginn \citep{geiping2025scaling}, all four tested models
with fixed input injection improve on both tasks at their deeper inference
settings.

After write-back, one-pass inference in LoopUS \citep{park2026loopus},
which uses gated recurrent updates, exceeds its original eight-pass accuracy
on both tasks.
Ouro \citep{zhu2025scaling}, which does not repeatedly inject a fixed input
representation, gains 12.51 and 11.04 points at one pass on GSM8K and GSM1K,
respectively, with lower four-pass accuracy on both tasks.

\subsection{Sources of depth-induced supervision}
\label{sec:ablations}

\begin{table}[tb]
\caption{\textbf{Useful self-distillation signals arise across teacher depths and trajectory
construction choices.}
Exact match (\%) with one-pass training. Panels vary the teacher depth, the question pool and the
target decoding (\emph{Setting}). The last two panels use the mean-depth-4 model, and all
other settings follow the default recipe. Changes are relative to the matching \emph{Mean depth} row.}
\label{tab:ablations}
\tablelayout
\resizebox{\textwidth}{!}{\begin{tabular}{@{}lcllllllc@{}}
\toprule
Model & Setting & \multicolumn{2}{c}{GSM8K} & \multicolumn{2}{c}{GSM1K} & \multicolumn{2}{c}{SVAMP} & Avg. \\
\cmidrule(lr){3-4} \cmidrule(lr){5-6} \cmidrule(lr){7-8}
&  & Depth 1 & Depth \(m\) & Depth 1 & Depth \(m\) & Depth 1 & Depth \(m\) & gain \\
\midrule
Mean depth 4 & -- & 40.86 & 56.18 & 36.93 & 51.12 & 52.40 & 63.30 & -- \\
\multirow{2}{*}{+ Self-distillation} & teacher depth 4 & 48.52 & \textbf{62.70}\,\dpos{+11.6\%} & 44.65 & \textbf{54.36}\,\dpos{+6.3\%} & 56.40 & \textbf{64.70}\,\dpos{+2.2\%} & \textcolor{deltapos}{+6.7\%} \\
 & teacher depth 1 & 38.06 & 59.29\,\dpos{+5.5\%} & 36.18 & 51.95\,\dpos{+1.6\%} & 53.10 & 63.60\,\dpos{+0.5\%} & \textcolor{deltapos}{+2.5\%} \\
\addlinespace[2pt]
Mean depth 32 & -- & 4.09 & 57.01 & 3.24 & 53.36 & 19.70 & \textbf{67.70} & -- \\
\multirow{2}{*}{+ Self-distillation} & teacher depth 32 & 31.84 & 62.09\,\dpos{+8.9\%} & 26.39 & \textbf{54.36}\,\dpos{+1.9\%} & 36.20 & 65.70\,\dneg{$-$3.0\%} & \textcolor{deltapos}{+2.6\%} \\
 & teacher depth 8 & 27.90 & \textbf{62.85}\,\dpos{+10.2\%} & 23.90 & 53.69\,\dpos{+0.6\%} & 34.70 & 65.10\,\dneg{$-$3.8\%} & \textcolor{deltapos}{+2.3\%} \\
\midrule
\multirow{5}{*}{+ Self-distillation} & GSM8K train (7,473) & 47.92 & 61.26\,\dpos{+9.0\%} & 40.66 & 52.95\,\dpos{+3.6\%} & 54.90 & 64.60\,\dpos{+2.1\%} & \textcolor{deltapos}{+4.9\%} \\
 & OpenMath (10,000) & 48.52 & 62.62\,\dpos{+11.5\%} & 42.16 & 53.94\,\dpos{+5.5\%} & 52.40 & 61.20\,\dneg{$-$3.3\%} & \textcolor{deltapos}{+4.6\%} \\
 & OpenMath, curated (10,000) & 46.78 & 61.56\,\dpos{+9.6\%} & 41.83 & 54.27\,\dpos{+6.2\%} & 54.10 & 63.90\,\dpos{+0.9\%} & \textcolor{deltapos}{+5.6\%} \\
 & OpenMath (30,000) & 50.11 & \textbf{63.38}\,\dpos{+12.8\%} & 43.49 & 53.61\,\dpos{+4.9\%} & 54.60 & 63.40\,\dpos{+0.2\%} & \textcolor{deltapos}{+5.9\%} \\
 & OpenMath (50,000) & 48.52 & 62.70\,\dpos{+11.6\%} & 44.65 & \textbf{54.36}\,\dpos{+6.3\%} & 56.40 & \textbf{64.70}\,\dpos{+2.2\%} & \textcolor{deltapos}{+6.7\%} \\
\midrule
\multirow{2}{*}{+ Self-distillation} & greedy & 48.52 & \textbf{62.62}\,\dpos{+11.5\%} & 42.16 & \textbf{53.94}\,\dpos{+5.5\%} & 52.40 & \textbf{61.20}\,\dneg{$-$3.3\%} & \textcolor{deltapos}{+4.6\%} \\
 & sampled, \(T{=}0.7\), \(k{=}10\) & 45.11 & 60.50\,\dpos{+7.7\%} & 41.24 & 53.03\,\dpos{+3.7\%} & 52.10 & 59.80\,\dneg{$-$5.5\%} & \textcolor{deltapos}{+2.0\%} \\
\bottomrule
\end{tabular}
}
\end{table}

\textbf{Useful supervision can emerge before the model reaches its standard
inference depth.} We compare teacher trajectories generated at different
depths while keeping one-pass learning and deeper inference fixed
(Table~\ref{tab:ablations}). For the mean-depth-32 model, learning from
eight-pass trajectories raises 32-pass GSM8K accuracy from 57.01\% to
62.85\%, compared with 62.09\% from 32-pass trajectories. Across the three
benchmarks, the corresponding mean per-task relative gains are 2.3\% and
2.6\%. Teacher depth can therefore be chosen separately from inference depth.

On the same 10,000 OpenMath questions, greedy continuations give higher
accuracy than truncated sampling on all three evaluated tasks, including
62.62\% versus 60.50\% on GSM8K (Table~\ref{tab:ablations}). Learning from
either GSM8K or OpenMath question pools also improves GSM8K and GSM1K in
the mean-depth-4 model. At a fixed pool size of 10,000 questions, selecting
questions for broader coverage raises SVAMP accuracy from 61.20\% with
random selection to 63.90\%, with both pools improving GSM8K.

\subsection{Retention of non-mathematical capabilities}
\label{sec:retention}

\begin{table}[!tbp]
\caption{\textbf{Average non-mathematical performance remains stable after write-back.}
OLMES averages at the mean depth over 17 tasks in six groups. Parenthesized
percentages show the average per-task relative change from the non-looped control
within each group. Per-task results are in
Table~\ref{tab:essd-retention}.}
\label{tab:retention}
\tablelayout
\resizebox{\textwidth}{!}{\begin{tabular}{@{}lccccccc@{}}
\toprule
Model & Knowledge (2) & Exams (2) & Commonsense (4) & Science QA (3) & Reading (3) & Open QA (3) & All (17) \\
\midrule
Non-looped control & 29.63 & 35.69 & 69.65 & 61.26 & 61.42 & 46.56 & 53.94 \\
\midrule
Mean-depth-4 Looped LM & 32.39\,\dchg{deltapos}{+9.0\%} & 39.97\,\dchg{deltapos}{+12.0\%} & 70.04\,\dchg{deltapos}{+0.6\%} & 66.01\,\dchg{deltapos}{+8.6\%} & 67.20\,\dchg{deltapos}{+10.5\%} & 47.45\,\dchg{deltapos}{+1.5\%} & 56.88\,\dchg{deltapos}{+6.3\%} \\
+ Self-distillation & 32.72\,\dchg{deltapos}{+10.5\%} & 40.07\,\dchg{deltapos}{+12.3\%} & 69.87\,\dchg{deltapos}{+0.3\%} & 65.91\,\dchg{deltapos}{+8.6\%} & 67.95\,\dchg{deltapos}{+12.3\%} & 47.59\,\dchg{deltapos}{+2.4\%} & 57.02\,\dchg{deltapos}{+6.9\%} \\
\midrule
Mean-depth-8 Looped LM & 34.34\,\dchg{deltapos}{+17.5\%} & 41.65\,\dchg{deltapos}{+16.7\%} & 69.91\,\dchg{deltapos}{+0.4\%} & 67.27\,\dchg{deltapos}{+10.8\%} & 64.94\,\dchg{deltapos}{+6.8\%} & 46.32\,\dchg{deltaneg}{$-$2.6\%} & 56.90\,\dchg{deltapos}{+6.8\%} \\
+ Self-distillation & 34.36\,\dchg{deltapos}{+17.0\%} & 41.62\,\dchg{deltapos}{+16.6\%} & 70.28\,\dchg{deltapos}{+1.0\%} & 67.47\,\dchg{deltapos}{+11.1\%} & 65.26\,\dchg{deltapos}{+7.7\%} & 46.43\,\dchg{deltaneg}{$-$2.1\%} & 57.09\,\dchg{deltapos}{+7.1\%} \\
\midrule
Mean-depth-32 Looped LM & 34.04\,\dchg{deltapos}{+16.5\%} & 41.86\,\dchg{deltapos}{+17.3\%} & 70.14\,\dchg{deltapos}{+0.8\%} & 67.02\,\dchg{deltapos}{+10.5\%} & 63.56\,\dchg{deltapos}{+3.1\%} & 45.89\,\dchg{deltaneg}{$-$3.3\%} & 56.58\,\dchg{deltapos}{+6.0\%} \\
+ Self-distillation & 33.79\,\dchg{deltapos}{+14.9\%} & 41.01\,\dchg{deltapos}{+14.9\%} & 70.18\,\dchg{deltapos}{+0.8\%} & 66.60\,\dchg{deltapos}{+9.8\%} & 63.18\,\dchg{deltapos}{+3.1\%} & 46.37\,\dchg{deltaneg}{$-$2.0\%} & 56.40\,\dchg{deltapos}{+5.6\%} \\
\bottomrule
\end{tabular}
}
\end{table}

\textbf{Write-back retains the overall non-mathematical performance of the
looped checkpoints.} We evaluate 17 OLMES tasks covering knowledge, exams,
commonsense reasoning, science QA, reading, and open-domain QA
(Table~\ref{tab:retention}). At their mean depths, the three looped checkpoints
score 56.58--56.90\% on the 17-task average, compared with 53.94\% for the
non-looped control. After learning from mathematical continuations, this
average changes by less than 0.2 percentage points for each model. The final
models achieve mean per-task relative gains of 5.6--7.1\% over the non-looped
control.

\section{Related Work}
\label{sec:related-work}

\paragraph{Recurrent depth and looped language models.}
Weight sharing across depth dates back to Universal Transformers \citep{dehghani2019universal}
and ALBERT \citep{lan2020albert}. Looped Transformers can emulate iterative algorithms
\citep{giannou2023looped} and approach the reasoning ability of deeper models
\citep{saunshi2025reasoning}. Recent work pre-trains recurrent-depth language models at scale
\citep{geiping2025scaling,zhu2025scaling}, converts pretrained models into looped form
\citep{bae2025relaxed,mcleish2025teaching,koishekenov2025encode}, and adapts the number of
passes to each token \citep{bae2025mixture} or to a compute budget \citep{jeddi2026loopformer}.
Building on these approaches, we develop looped language models through recurrent mid-training
and study how the predictive behavior revealed by additional passes can inform subsequent training.

\paragraph{Self-distillation and model-generated supervision.}
Extra inference-time computation can improve reasoning \citep{snell2025scaling}.
Several lines of work use model predictions as supervision, including sequence-level distillation
\citep{kim2016sequence} and born-again self-distillation \citep{furlanello2018born}.
Related work internalizes chain-of-thought into latent reasoning \citep{hao2025training,shen2025codi}
and trains on a model's own generations, often filtered for correctness or consistency
\citep{zelikman2022star,huang2023large,singh2024beyond} and sometimes without any filter
\citep{zhang2026ssd,agarwal2025unreasonable}. We use sequence-level self-distillation
on the model's own deeper continuations to update only the shared recurrent core.

\paragraph{Supervision across depth and training through partial computation.}
Intermediate depths are often supervised directly through losses on hidden layers
\citep{lee2015deeply}, deep-to-shallow self-distillation \citep{zhang2019your} or early-exit
losses \citep{elhoushi2024layerskip}. Weight-tied iterative models are often trained through part
of their computation. Implicit models use implicit or approximate gradients, including
approximations based on a single operator application \citep{bai2019deep,geng2021training,fung2022jfb}.
Recurrent networks that extrapolate to more iterations \citep{schwarzschild2021can} have been
trained by backpropagating through a random suffix \citep{bansal2022endtoend}.
Recurrent-depth language models also truncate backpropagation \citep{geiping2025scaling},
as in truncated backpropagation through time \citep{williams1990efficient}.
Preference write-back separates the forward depth used for learning from the depths used to
generate supervision and to run inference. Our pass-wise intervention shows that an update
learned through one pass remains useful at later recurrent passes. Controlled comparisons
further show that the advantage of single-pass learning over full-depth learning depends on
the recurrent recipe.

\section{Conclusion}
\label{sec:conclusion}

Once a model has learned to benefit from recurrent depth, its own
deeper executions can provide supervision for further learning.
Our analysis shows how additional recurrence refines predictive
preferences among already plausible tokens, and preference write-back
demonstrates how the model can learn from these changes to improve
its reasoning.
An important finding is that an update learned through a single
application of the shared recurrent operator can improve its
subsequent applications. Using the updated core only at later
passes recovers nearly the full gain in our four-pass GSM8K experiment.
The advantage of shallow learning over full-depth learning also
varies with the recurrent recipe, linking the development of
recurrent computation to its subsequent improvement.
More broadly, these findings open up the question of how the behavior
that emerges as recurrent computation develops can guide its
continued learning.

\bibliography{references}
\bibliographystyle{plainnat}

\appendix
\section{Recurrence during mid-training}
\label{sec:app-midtraining-progress}

\begin{figure}[ht]
\centering
\includegraphics[width=\linewidth]{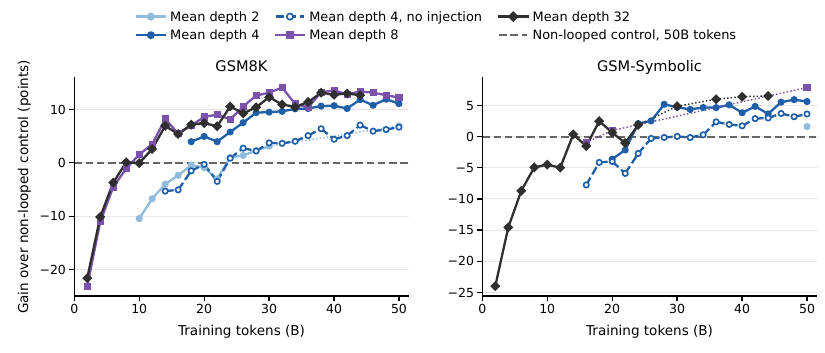}
\caption{Gain in exact match (points) over the non-looped control after 50B tokens (dashed line
at zero) during mid-training, for the Looped LMs of this paper, each evaluated at a fixed depth
(its mean depth; eight passes for the mean-depth-32 model). Dotted segments join checkpoints
more than 4B tokens apart.}
\label{fig:app-midtraining-progress}
\end{figure}

Before asking what additional recurrence contributes to a trained model, we check that
recurrence pays off during mid-training itself. Figure~\ref{fig:app-midtraining-progress}
follows the Looped LMs of this paper on GSM8K and GSM-Symbolic as mid-training proceeds from
the stage-1 checkpoint (Appendix~\ref{sec:app-midtraining}), as their gain over the non-looped
control trained on the same 50B tokens. Every Looped LM finishes with a positive gain on both
tasks, and on GSM8K each one crosses zero within 24B tokens, less than half of the control's
training.
Recurrent computation is therefore already an effective part of these models before any
self-distillation; the remaining appendices and Sections~\ref{sec:method}
and~\ref{sec:experiments} study what it adds and how to write it back into the shared block.

\FloatBarrier

\section{Recurrence primarily reweights the top of the distribution}
\label{sec:stable-decision-probability}

\begin{figure}[tb]
\centering
\includegraphics[width=\linewidth]{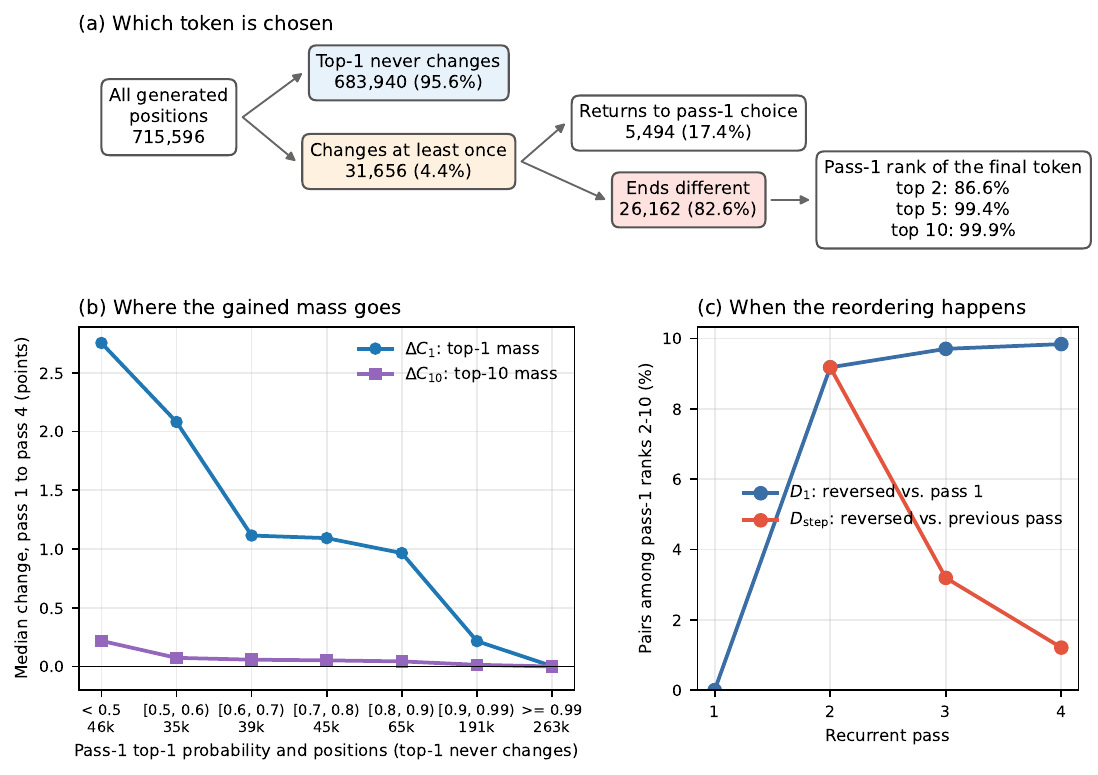}
\caption{Effect of passes 2--4 on the readout distribution over 715,596 generated positions
of GSM8K training problems. (a) Stability of the top-1 token across passes, and the pass-1
rank of the final token when it differs; each percentage is relative to the box it branches
from. (b) Median change from \(q_1\) to \(q_4\) in top-1 probability (\(\Delta C_1\)) and
top-10 mass (\(\Delta C_{10}\)), binned by the pass-1 top-1 probability, at positions whose
top-1 never changes. (c) Pairwise reordering among the
pass-1 ranks 2--10, relative to pass 1 (\(D_{1}\)) and to the preceding pass
(\(D_{\mathrm{step}}\)).}
\label{fig:stable-decision-summary}
\end{figure}

\paragraph{Reading out along a depth-4 trajectory.}
We analyze the mean-depth-4 Looped LM of Table~\ref{tab:main}. During an ordinary greedy
generation at depth 4, we additionally decode the state \(s_k\) after each pass
\(k = 1,\ldots,4\) with the same output stack and language-model head, giving the
intermediate readouts
\[
q_k(\cdot \mid x, \hat y_{<t}) = \operatorname{softmax}\!\left(W[C(s_k)]_n\right),
\qquad k = 1,\ldots,4 .
\]
The readouts are diagnostic only: each decodes from a copy of the output-stack cache, none
feeds back into the recurrence, and tokens are selected from \(q_4\) alone, so the generation
is exactly the ordinary depth-4 generation. Because every \(q_k\) is conditioned on the
depth-4 trajectory, and its output stack attends to the depth-4 states of earlier positions, it
describes an intermediate state of that execution and differs in general from \(p_k\), the distribution of an independent execution of depth \(k\)
(Section~\ref{sec:looped-lm}). Figure~\ref{fig:stable-decision-summary} summarizes how the
readouts change from pass to pass.

\paragraph{Recurrence primarily reshapes the head of the distribution.}
Additional passes mainly reshape the relative probabilities of candidates already near the
top after the first pass, rather than promote previously low-ranked candidates. The effect
varies in magnitude: at many positions the probabilities change while the ordering is
preserved, at others nearby candidates are reordered, and only occasionally does the top-1
prediction change.

We apply this diagnostic to all 7,473 GSM8K training problems under greedy depth-4
generation, yielding 715,596 generated-token positions.
\begin{enumerate*}[label=(\alph*)]
    \item The top-1 prediction is stable at most positions: it is the same at all four
    passes at 95.6\% of positions, and when the final top-1 token differs from the pass-1
    choice, it was already among the pass-1 top five 99.4\% of the time
    (Figure~\ref{fig:stable-decision-summary}a);
    \item Where the top-1 remains fixed, recurrence mainly redistributes probability within
    the head of the distribution: the top-1 probability increases while the total top-10
    mass changes little, with the largest shifts where the first pass is uncertain and the
    effect vanishing as it becomes confident (Figure~\ref{fig:stable-decision-summary}b);
    \item When recurrence does reorder nearby candidates, most of that change occurs early:
    pass 2 already accounts for 93\% of the pairwise reordering among the pass-1 ranks 2--10
    observed by pass 4, while later passes make progressively smaller changes
    (Figure~\ref{fig:stable-decision-summary}c).
\end{enumerate*}

\paragraph{From readout to supervision.}
Within the four-pass generation, additional passes mainly redistribute probability
among tokens that already rank highly in the first-pass readout of that same trajectory.
Greedy decoding at the deeper execution turns this preference into a complete continuation,
and these continuations are the self-distillation targets of
Section~\ref{sec:depth-supervision}.

\paragraph{Relation to layer-wise readouts.}
Decoding intermediate states with the model's own output head is a common probe of residual
networks, whose layers refine a prediction iteratively \citep{jastrzebski2018residual}:
feed-forward updates promote candidate tokens in vocabulary space
\citep{geva2022transformer}, late layers sharpen the next-token distribution
\citep{lad2025remarkable}, and intermediate predictions can be more accurate than final ones
\citep{halawi2024overthinking}. In a looped model every pass shares one state space and one
operator, so the same output stack reads out each pass directly, without learned per-layer
probes.

\section{Implementation details}
\label{sec:experimental-details}

This appendix gives the implementation behind Sections~\ref{sec:method}
and~\ref{sec:experiments}: the looped architecture
(Appendix~\ref{sec:app-architecture}), the mid-training that produces each Looped LM
(Appendix~\ref{sec:app-midtraining}), the self-distillation recipe
(Appendix~\ref{sec:app-self-distillation}), and the evaluation protocol
(Appendix~\ref{sec:app-evaluation}).

\paragraph{Notation in the tables.}
The tables name the three depths of Section~\ref{sec:method-writeback} directly:
\emph{Teacher} is the teacher depth \(r_{\mathrm T}\), \emph{Training} is the training depth
\(r_{\mathrm L}\), and the \emph{Depth 1} and \emph{Mean depth} columns give the inference depth
\(r_{\mathrm E}\). The mean depth \(m\) is the target of the depth curriculum during
mid-training (Appendix~\ref{sec:app-midtraining}), and it is also the inference depth at which
each Looped LM is used.

\subsection{Architecture}
\label{sec:app-architecture}

\paragraph{Partition of OLMo-2 1B.}
OLMo-2 1B has 16 Transformer blocks of width \(d=2048\) with 16 attention heads
\citep{olmo2}; its released configuration specifies a SwiGLU feed-forward layer of width 8,192
and a vocabulary of 100,352 tokens. The input stack \(P\) consists of blocks 1--7, the shared recurrent core
\(F_\theta\), called the shared recurrent block in Section~\ref{sec:experiments}, of blocks
8--11, and the output stack \(C\) of blocks 12--16.
\(F_\theta\) is a single module invoked once per recurrent pass, so an execution of depth
\(r\) applies the same 268,468,224 parameters \(r\) times; the complete model has
1.49B parameters.

\paragraph{Input injection and the initial state.}
The injection projection \(A_\psi\) is one bias-free linear map from
\(\mathbb{R}^{2d}\) to \(\mathbb{R}^{d}\), applied at each position, with no gate,
normalization or activation. The same input-stack output \(e\) is concatenated with the
recurrent state at every pass. During training, the initial state \(s_0\) is drawn from a
truncated normal distribution with standard deviation \(0.0087\). At evaluation, \(s_0\) is
drawn with a seed derived from the prompt tokens, so executions of different depth on the
same prompt start from the same state.

\paragraph{Model without input injection.}
The model without input injection in Table~\ref{tab:regime} removes both the
concatenation and the separate initial state:
\[
s_0 = e, \qquad s_i = F_\theta(s_{i-1}), \qquad i = 1,\ldots,r .
\]
It is therefore a different recurrence rather than injection with a zeroed projection.
Its partition, output stack and mid-training procedure match those of the injected
mean-depth-4 model.

\subsection{Mid-training of the Looped LMs}
\label{sec:app-midtraining}

\paragraph{Common setup.}
Every Looped LM and the non-looped control start from the OLMo-2 1B stage-1 checkpoint
and continue causal language-model training on the Dolmino mixture \citep{olmo2} for up
to 50B tokens, with context length 4,096, a global batch of 2,097,152 tokens (512
sequences), the same data order, BF16 precision and fully sharded data parallelism.
Table~\ref{tab:app-midtraining} lists the optimizer and the gradient path of each model. All
models use a peak learning rate of \(7.4\times10^{-5}\), decayed linearly to zero over 23,842
steps without warmup, \(\beta = (0.9, 0.95)\), weight decay 0.033, gradient clipping at norm
1.0 and a z-loss coefficient of \(10^{-5}\). Muon updates the matrix parameters with its
learning rate rescaled to the update scale of AdamW, and AdamW updates the embeddings,
normalization parameters and language-model head.

\begin{table}[tbp]
\caption{Mid-training of the Looped LMs and the non-looped control. \emph{Backpropagation}
gives the passes that carry gradient at each training step.}
\label{tab:app-midtraining}
\tablelayout
\small
\begin{tabular*}{0.84\linewidth}{@{\extracolsep{\fill}}lccl@{}}
\toprule
Model & Input injection & Optimizer & Backpropagation \\
\midrule
Non-looped control & \no & AdamW & single fixed pass \\
Mean-depth-2 Looped LM & \yes & Muon & every pass \\
Mean-depth-4 Looped LM & \yes & Muon & every pass \\
Mean-depth-4 Looped LM, no injection & \no & Muon & every pass \\
Mean-depth-8 Looped LM & \yes & AdamW & last eight passes \\
Mean-depth-32 Looped LM & \yes & AdamW & last eight passes \\
\bottomrule
\end{tabular*}
\end{table}

\paragraph{Depth sampling and curriculum.}
At each training step, one depth \(r\) is drawn for the whole batch from the
Poisson-lognormal distribution of \citet{geiping2025scaling}, with log-normal scale
\(\sigma = 0.5\) and parameterized by the current curriculum mean \(\mu_t\) of
Section~\ref{sec:looped-lm}. Over the first 75\% of the \(T\) training steps, \(\mu_t\) rises
from 1 to the target mean depth \(m\) along a one-minus-square-root schedule, and it is held
at \(m\) afterwards:
\[
\mu_t = \max\!\left(1,\;
\left\lceil m\left(1 - \sqrt{1 - \min(t/t_{\mathrm w},\,1)}\right)\right\rceil\right),
\qquad t_{\mathrm w} = 0.75\,T .
\]

\paragraph{Gradient path through the recurrence.}
For the mean-depth-8 and mean-depth-32 models, all passes before the last eight of a
sampled depth run without gradient, so backpropagation covers at most eight passes,
following \citet{geiping2025scaling}; this bounds the length of the gradient path through
the recurrence and keeps optimization stable at large depth. Because \(e\) is injected at
every pass, the input stack still receives gradient through the differentiated passes.

\paragraph{Non-looped control.}
The non-looped control of Tables~\ref{tab:main} and~\ref{tab:retention} is trained with the
same data and setup, but with a single fixed recurrent pass and no input injection. With
\(s_0 = e\) and one pass, it computes \(C(F_\theta(P(x)))\), which is the original
16-block OLMo-2 architecture. Appendix~\ref{sec:app-midtraining-progress} follows the accuracy
of the Looped LMs through this stage.

\subsection{Self-distillation}
\label{sec:app-self-distillation}

\paragraph{Teacher trajectories.}
The teacher is the frozen Looped LM before the update. For each training question it
generates one continuation at teacher depth \(r_{\mathrm T}\) by greedy decoding, using the
fixed eight-shot chain-of-thought prompt of OLMES for GSM8K \citep{gu2025olmes,olmo2} and at most 512
new tokens. Generation ends at the first stop string or end-of-sequence token. The
generated token ids, including the stop token, are the target without further editing; a
continuation that reaches the token limit is kept as generated. Targets are generated once
and reused in every epoch, and the reference answer of the training question is never used.
The sampled targets in Table~\ref{tab:ablations} apply temperature \(T=0.7\) and then
top-\(k\) truncation with \(k=10\) at each step.

\paragraph{Objective and updated parameters.}
A training sequence is the eight-shot prompt followed by the target continuation. The loss
is the completion-only next-token cross-entropy of Section~\ref{sec:method-writeback}, normalized by
the total number of target tokens in the batch; no distillation term on the teacher
probabilities is added. Only the 32 parameter tensors of \(F_\theta\) are updated. The
embeddings, input stack, injection projection \(A_\psi\), output stack and language-model
head are frozen, leaving 18\% of the parameters trainable.

Algorithm~\ref{alg:writeback} summarizes the procedure.

\begin{algorithm}[tbp]
\caption{Preference write-back by self-distillation}
\label{alg:writeback}
\small
\begin{tabbing}
\hspace{1.3em}\=\hspace{1.3em}\=\hspace{1.3em}\=\kill
\textbf{Input:} Looped LM with shared core \(\theta_0\) and remaining parameters \(\phi\);
prompts \(\{x^{(j)}\}_{j=1}^{N}\);\\
\>teacher depth \(r_{\mathrm T}\); training depth \(r_{\mathrm L}\); number of epochs \(E\)\\[3pt]
\textit{Expose: the frozen model generates targets at teacher depth}\\
\textbf{for} \(j = 1,\ldots,N\) \textbf{do}\\
\>\(\hat y^{(j)} \leftarrow\) greedy continuation of \(x^{(j)}\) under
\(p^{(0)}_{r_{\mathrm T}}\), at most 512 tokens, kept unfiltered\\
\(\mathcal D_{r_{\mathrm T}} \leftarrow \{(x^{(j)}, \hat y^{(j)})\}_{j=1}^{N}\)\\[3pt]
\textit{Incorporate: update only the shared core through \(r_{\mathrm L}\) passes}\\
\(\theta \leftarrow \theta_0\)\\
\textbf{for} epoch \(= 1,\ldots,E\) \textbf{do}\\
\>\textbf{for} each batch \(B \subset \mathcal D_{r_{\mathrm T}}\) of 16 sequences \textbf{do}\\
\>\>\(\theta \leftarrow \operatorname{AdamW}\big(\theta, \nabla_\theta
\mathcal L_{r_{\mathrm L}}(\theta; B)\big)\)
\quad \textit{completion-only cross-entropy; \(\phi\) frozen}\\[3pt]
\textit{Reuse: the updated core serves every pass at any inference depth}\\
\textbf{return} \(\theta_1 \leftarrow \theta\), used with \(\phi\) at inference depth
\(r_{\mathrm E}\)
\end{tabbing}
\end{algorithm}

\paragraph{Training depth.}
Training at depth \(r_{\mathrm L}\) runs \(r_{\mathrm L}\) recurrent passes forward, reads
the loss out after the last pass, and backpropagates through all \(r_{\mathrm L}\) passes;
because \(F_\theta\) is shared, the gradient accumulates contributions from each of its
uses. The \(4^\dagger\) row of Table~\ref{tab:mechanism} runs the first three passes
without gradient and backpropagates through the fourth only. For the depth-8 and depth-32
training rows of Table~\ref{tab:mechanism}, activations are recomputed pass by pass to fit
in memory, which leaves the gradient unchanged.

\paragraph{Optimization.}
Table~\ref{tab:app-sd-hparams} lists the optimization settings, which are shared by every
self-distillation run in the paper. Runs differ only in the teacher depth, training depth,
question pool and target decoding stated in the corresponding table.

\begin{table}[tbp]
\caption{Self-distillation settings shared by all runs, and the size of the OpenMath
question pools.}
\label{tab:app-sd-hparams}
\tablelayout
\small
\begin{tabular*}{0.84\linewidth}{@{\extracolsep{\fill}}ll@{}}
\toprule
Setting & Value \\
\midrule
Updated parameters & shared recurrent block \(F_\theta\) (268M of 1.49B) \\
Objective & completion-only next-token cross-entropy \\
Optimizer & AdamW, \(\beta = (0.9, 0.95)\), no weight decay \\
Learning rate & \(3\times10^{-6}\); 10 linear warmup steps, then cosine decay \\
Batch & 16 sequences \\
Gradient clipping & global norm 1.0 \\
Epochs & 3 (OpenMath pools), 6 (GSM8K training questions) \\
Precision & BF16 computation, FP32 parameters \\
Targets & one teacher continuation per question, at most 512 tokens \\
\midrule
\multicolumn{2}{@{}l}{\emph{OpenMath question pools} {\scriptsize(targets of the mean-depth-4 model)}} \\
10,000 questions & 992,459 target tokens per epoch \\
10,000 questions, curated & 1,019,832 target tokens per epoch \\
30,000 questions & 2,973,623 target tokens per epoch \\
50,000 questions & 4,953,856 target tokens per epoch \\
\bottomrule
\end{tabular*}
\end{table}

\paragraph{Question pools.}
The default pool contains question text from the \texttt{augmented\_gsm8k} portion of
OpenMathInstruct-2 \citep{toshniwal2025openmathinstruct}. Only the problem statements are
read; the released solutions and answers are not used. Questions are normalized for
Unicode and whitespace, deduplicated, and screened against the GSM8K test, GSM1K and SVAMP
questions and the development questions of Appendix~\ref{sec:app-evaluation}. A lexical audit
against these 4,214 questions (word five-gram Jaccard similarity of at least 0.8, or
numeric-template similarity of at least 0.9) finds no near match in the final pools. The 10,000-, 30,000- and 50,000-question pools
are nested. The curated 10,000-question pool of Table~\ref{tab:ablations} is selected from
the 50,000-question pool using the question text alone: it matches the length distribution of
the random 10,000-question pool, favors rare problem types, and keeps at most one question
from each cluster of near-duplicates, and it reuses the targets already generated for those
questions. The GSM8K pool consists of all 7,473 training questions \citep{cobbe2021gsm8k}.

\paragraph{Checkpoint selection.}
For the OpenMath runs in Table~\ref{tab:main}, we selected checkpoints by
development-set accuracy at each model's mean depth. For the full GSM8K training
run in Table~\ref{tab:ablations}, we selected epoch 2 by mean relative gain in
test accuracy at four passes.

\subsection{Evaluation protocol}
\label{sec:app-evaluation}

\paragraph{Mathematical reasoning.}
Table~\ref{tab:app-eval-sets} lists the mathematical evaluation sets. Every model is
evaluated with greedy decoding in BF16 at depth 1 and at its mean depth.
Generation stops at the first stop string or end-of-sequence token, and a response that
reaches the token limit remains in the denominator. For GSM8K, GSM1K and SVAMP the answer
is the number after the last \texttt{\#\#\#\#} if present and otherwise the last number in the
response, normalized for commas, trailing zeros and sign; GSM-Symbolic uses the answer
extraction of the OLMES GSM8K task; for ASDiv it is the number after ``The final answer is'',
falling back to the last number; for AQuA-RAT it is the last explicitly labelled option letter.
Accuracy is exact match of the extracted answer with the reference answer.

\begin{table}[tbp]
\caption{Mathematical evaluation sets. All prompts are fixed chain-of-thought
demonstrations shared by every question of a set.}
\label{tab:app-eval-sets}
\tablelayout
\small
\begin{tabular*}{0.84\linewidth}{@{\extracolsep{\fill}}lrcr@{}}
\toprule
Benchmark & Questions & Shots & Token limit \\
\midrule
GSM8K test \citep{cobbe2021gsm8k} & 1,319 & 8 & 512 \\
GSM1K \citep{zhang2024gsm1k} & 1,205 & 8 & 512 \\
SVAMP \citep{patel2021svamp} & 1,000 & 8 & 512 \\
GSM-Symbolic, main \citep{mirzadeh2025gsmsymbolic} & 5,000 & 8 & 1,024 \\
ASDiv \citep{miao2020asdiv} & 2,305 & 8 & 512 \\
AQuA-RAT \citep{ling2017aqua} & 254 & 4 & 512 \\
\bottomrule
\end{tabular*}
\end{table}

GSM8K and GSM-Symbolic use the OLMES eight-shot prompt \citep{gu2025olmes,olmo2}; GSM1K and SVAMP
use the eight arithmetic chain-of-thought demonstrations of \citet{wei2022chain}, and ASDiv
uses the same eight demonstrations in a problem--solution format. AQuA-RAT uses four fixed
demonstrations with their rationales, drawn from its development split. The
non-mathematical tasks of Tables~\ref{tab:retention} and~\ref{tab:essd-retention} are
evaluated with the OLMES harness at each model's mean depth (depth 1 for the non-looped
control), with the shots and metrics listed in Table~\ref{tab:essd-retention}.

\paragraph{Summary columns.}
Relative changes in the tables are per-task changes divided by the reference value, the Looped
LM before self-distillation or, where stated, the non-looped control. The \emph{Avg.} columns
of Tables~\ref{tab:main} and~\ref{tab:mechanism}--\ref{tab:ablations} are unweighted means of
these per-task relative changes. In Table~\ref{tab:retention}, each group score is the mean of
its task scores and its percentage the mean of the per-task relative changes.

\paragraph{Development set and intervals.}
We used a fixed development set of 690 questions from the full GSM8K training
pool and excluded those questions from the OpenMath pools.
Intervals are 95\% question-paired bootstrap intervals over 10,000 resamples of the evaluation
questions. Table~\ref{tab:app-paired} gives them for the comparisons of Table~\ref{tab:main}:
every Looped LM is above the non-looped control on all six tasks, and the gains of
self-distillation on GSM8K and GSM-Symbolic exclude zero for all three models.

\begin{table}[tbp]
\caption{Paired differences in exact match (points) behind Table~\ref{tab:main}, at each
model's mean depth, with 95\% question-paired bootstrap intervals. The non-looped control is
evaluated at depth 1.}
\label{tab:app-paired}
\tablelayout
\footnotesize
\resizebox{\linewidth}{!}{\begin{tabular}{@{}lcccccc@{}}
\toprule
Comparison & GSM8K & GSM1K & SVAMP & GSM-Symbolic & ASDiv & AQuA-RAT \\
\midrule
\multicolumn{7}{@{}l}{\emph{+ Self-distillation minus Looped LM}} \\
Mean depth 4 & \(+6.5\) [4.2, 8.8] & \(+3.2\) [1.1, 5.5] & \(+1.4\) [\(-1.1\), 3.9] & \(+4.1\) [3.0, 5.3] & \(+1.7\) [0.5, 3.0] & \(+1.2\) [\(-3.9\), 6.3] \\
Mean depth 8 & \(+6.5\) [4.4, 8.6] & \(+3.5\) [1.2, 5.8] & \(-0.8\) [\(-3.0\), 1.4] & \(+3.4\) [2.3, 4.5] & \(+1.7\) [0.5, 3.0] & \(-1.2\) [\(-6.7\), 4.3] \\
Mean depth 32 & \(+5.1\) [2.7, 7.6] & \(+1.0\) [\(-1.6\), 3.6] & \(-2.0\) [\(-4.6\), 0.6] & \(+3.2\) [1.9, 4.4] & -- & \(+5.5\) [0.0, 11.0] \\
\midrule
\multicolumn{7}{@{}l}{\emph{Looped LM minus non-looped control}} \\
Mean depth 4 & \(+12.1\) [9.6, 14.6] & \(+11.3\) [8.7, 13.8] & \(+6.8\) [4.0, 9.7] & \(+6.1\) [4.9, 7.3] & \(+5.4\) [3.9, 6.9] & \(+6.7\) [2.0, 11.8] \\
Mean depth 8 & \(+12.3\) [9.8, 14.8] & \(+12.7\) [10.2, 15.2] & \(+9.5\) [6.7, 12.3] & \(+7.6\) [6.4, 8.9] & \(+6.8\) [5.3, 8.3] & \(+13.4\) [7.9, 19.3] \\
Mean depth 32 & \(+13.0\) [10.4, 15.5] & \(+13.5\) [11.0, 16.2] & \(+11.2\) [8.3, 14.2] & \(+7.5\) [6.2, 8.7] & \(+7.2\) [5.7, 8.7] & \(+9.1\) [3.5, 15.0] \\
\bottomrule
\end{tabular}}
\end{table}

\subsection{Compute}
\label{sec:app-compute}

Mid-training runs on NVIDIA A100 80GB GPUs, and self-distillation, including the generation
of teacher trajectories, on NVIDIA A800 80GB GPUs. Self-distillation is small next to
mid-training. Even at teacher depth 32, generating the trajectories for 10,000 questions takes
18 GPU-hours, and three epochs of training on them at depth 1 take 2 GPU-hours.

\section{Further evidence on training depth}
\label{sec:app-training-depth}

This appendix supports Sections~\ref{sec:training-depth} and~\ref{sec:core-position}. It
details the gradient measurements of Figure~\ref{fig:gradient-geometry}
(Appendix~\ref{sec:gradient-diagnostic}), follows the two training depths across inference
depths and model families (Appendix~\ref{sec:app-depth-profile}), and gives
the full record of the pass-wise block replacement of Table~\ref{tab:core-position}
(Appendix~\ref{sec:app-core-replacement}).

\subsection{Recurrent pass-gradient diagnostic}
\label{sec:gradient-diagnostic}

\paragraph{Setup.}
Both panels of Figure~\ref{fig:gradient-geometry} use the mean-depth-4 Looped LM before
self-distillation, the point from which training at depth 1 and training at depth 4 start
with the same parameters and the same targets. We take 64 fixed teacher trajectories from
the OpenMath question pool (6,892 target tokens) and evaluate the completion-only loss of
Section~\ref{sec:method-writeback} in FP32.

\paragraph{Panel (a): gradient at the same first-pass state.}
For each trajectory, let \(s_1\) be the state after the first recurrent pass and \(L_k\) the
loss read out after \(k\) passes. We compute \(g_k = \partial L_k / \partial s_1\) over the
full state tensor for \(k = 1,\ldots,4\); the state \(s_1\) is verified to be identical
across the four depths. The panel plots the cosine between \(g_k\) and the one-pass
gradient \(g_1\), and the ratio \(\lVert g_k\rVert / \lVert g_1\rVert\), each computed per
trajectory and averaged over the 64 trajectories; error bars are 95\% bootstrap intervals
over trajectories. Most of the change occurs at the first additional pass: with two passes,
the cosine is already 0.10 and the norm 32\% of the one-pass gradient.

\paragraph{Panel (b): parameter gradient by pass.}
To attribute the gradient of four-pass training to individual uses of the shared block,
we untie the block into four copies with identical weights, one per pass. Their gradients
\(g^{(1)},\ldots,g^{(4)}\) sum to the gradient of the shared parameters, with a maximum
relative error of \(1.3\times10^{-6}\). The panel shows the share
\(\lVert g^{(i)}\rVert / \sum_j \lVert g^{(j)}\rVert\) for each pass, computed on four groups
of 16 trajectories, the batch size of self-distillation, and averaged over the groups.

Together, the two panels suggest that training at depth 1 applies the direct signal to the first
use of the shared block, whereas training at depth 4 reaches it only as a small, rotated
remnant and updates the block mainly through its uses on late recurrent states.

\subsection{Training depth across inference depths and models}
\label{sec:app-depth-profile}

\begin{figure}[!htb]
\centering
\includegraphics[width=\linewidth]{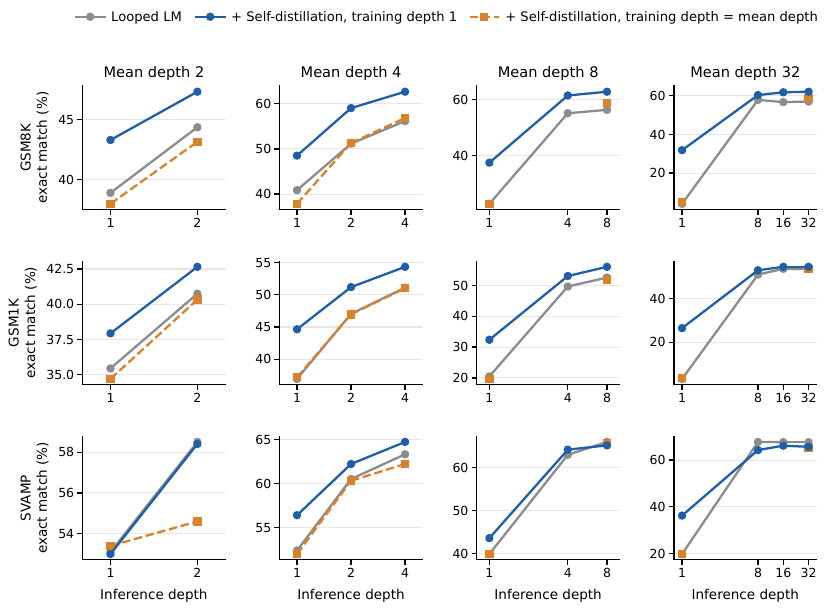}
\caption{Exact match (\%) across inference depths for four injected Looped LMs (columns) on
GSM8K, GSM1K and SVAMP (rows). Each Looped LM is self-distilled from its own greedy
mean-depth trajectories through one pass or through its full mean depth; lines connect
adjacent evaluated depths. Inference depth is on a logarithmic axis, and the vertical range
differs between panels. The values at depth 1 and at the mean depth of the mean-depth-4,
8 and 32 models are those of Table~\ref{tab:mechanism}.}
\label{fig:app-depth-profile}
\end{figure}

Figure~\ref{fig:app-depth-profile} follows each injected Looped LM and its two self-distilled
models across the inference depths we evaluated, for mean depths 2, 4, 8 and 32. On GSM8K
and GSM1K, training at depth 1 improves the model at every inference depth in all four
families, and the gain from recurrence is retained: the mean-depth-4 model self-distilled at
depth 1 rises from 48.52\% on GSM8K at one pass to 59.06\% at two and 62.70\% at four. The
gain at one pass is largest where that execution starts weakest: the mean-depth-32 model moves
from 4.09\% to 31.84\% at one pass while also improving at 8, 16 and 32 passes. Training on
the same targets through the full mean depth ends below training at depth 1 at the mean depth
on both sets in every family.

The mean-depth-2 model repeats the controlled comparison of Table~\ref{tab:mechanism} on a
second injected checkpoint, mid-trained toward mean depth 2: the teacher generates one greedy
trajectory at depth 2 for each of the 10,000 OpenMath questions, and the same targets train
the shared block through one pass or through two, with all other settings as in
Appendix~\ref{sec:app-self-distillation}. At the deployed depth, training at depth 1 is ahead
by 4.2, 2.3 and 3.8 points on GSM8K, GSM1K and SVAMP, and every 95\% CI of these paired
differences excludes zero. Training at depth 1 is also the cheaper update: on the same
trajectories and hardware, three epochs of the mean-depth-4 model take 0.72 GPU-hours at
depth 1 against 1.14 at depth 4, with 14\% less peak memory.

\subsection{Pass-wise block replacement}
\label{sec:app-core-replacement}

The intervention of Table~\ref{tab:core-position} executes the mean-depth-4 Looped LM for
four passes and chooses, for each pass, either the original shared block or the block of the
self-distilled model of Table~\ref{tab:main}. All other parameters are shared by the two
models; only the 32 tensors of the recurrent block differ. Each pattern is a complete
execution: generation and the cache are rebuilt under that pattern, so the comparison locates
where the updated block is needed rather than splitting the gain into additive per-pass
contributions. On eight check examples, the two pure patterns reproduce ordinary inference of the two
models exactly, in generated tokens, next-token logits and cache contents.

Table~\ref{tab:app-core-replacement} gives the record of the four patterns on GSM8K. Using the updated
block on passes 2--4 recovers 97\% of the gain obtained by using it throughout. Using it only
on the first pass corrected 39 answers and introduced 38 errors, for a net gain of
one correct answer out of 1,319.

\begin{table}[tbp]
\caption{Pass-wise replacement of the shared block on GSM8K (1,319 questions) at four-pass
inference. \yes: the self-distilled block is used at that pass; \no: the original block.
Change is in points relative to the original model; repaired / broken count the questions
that become correct / incorrect.}
\label{tab:app-core-replacement}
\tablelayout
\small
\begin{tabular*}{0.84\linewidth}{@{\extracolsep{\fill}}cccccccc@{}}
\toprule
\multicolumn{4}{c}{Pass} & & & & \\
\cmidrule(lr){1-4}
1 & 2 & 3 & 4 & Accuracy (\%) & Change & 95\% CI & Repaired / broken \\
\midrule
\no & \no & \no & \no & 56.18 & -- & -- & -- \\
\yes & \no & \no & \no & 56.25 & \(+0.1\) & [\(-1.2\), 1.4] & 39 / 38 \\
\no & \yes & \yes & \yes & 62.47 & \(+6.3\) & [4.0, 8.6] & 164 / 81 \\
\yes & \yes & \yes & \yes & 62.70 & \(+6.5\) & [4.2, 8.8] & 165 / 79 \\
\bottomrule
\end{tabular*}
\end{table}

\FloatBarrier

\section{Retention beyond mathematics}
\label{sec:app-retention}

Table~\ref{tab:essd-retention} gives the per-task scores behind Table~\ref{tab:retention}.
Self-distillation leaves individual tasks close to where they were: 47 of the 51 task scores
of the three self-distilled models move by less than one point, and the mean-depth-8 model
stays within one point on every task. The advantage that looping gives over the non-looped
control is preserved as well: after self-distillation, each Looped LM remains above the
control on at least 13 of the 17 tasks.

\begin{table}[H]
\caption{Scores (\%) on the 17 non-mathematical OLMES tasks, grouped as in
Table~\ref{tab:retention}: the non-looped control at depth 1 and, for each Looped LM, the model
before and after self-distillation at its mean depth. The self-distilled models are those of
Table~\ref{tab:main}; shots and metrics follow the OLMES task definitions
(Appendix~\ref{sec:app-evaluation}), and AGIEval is evaluated in both its 3-shot and its 1-shot
configuration.}
\label{tab:essd-retention}
\tablelayout
\footnotesize
\resizebox{\linewidth}{!}{\begin{tabular}{@{}lclrrrrrrr@{}}
\toprule
& & & Control & \multicolumn{2}{c}{Mean-depth-4} & \multicolumn{2}{c}{Mean-depth-8} & \multicolumn{2}{c}{Mean-depth-32} \\
\cmidrule(lr){5-6} \cmidrule(lr){7-8} \cmidrule(lr){9-10}
Task & Shots & Metric & Depth 1 & Looped LM & + Self-dist. & Looped LM & + Self-dist. & Looped LM & + Self-dist. \\
\midrule
\multicolumn{10}{@{}l}{\emph{Knowledge}} \\
MMLU & 5 & accuracy & 43.92 & 48.17 & 48.44 & 50.14 & 50.43 & 49.69 & 49.66 \\
MMLU-Pro & 5 & accuracy & 15.33 & 16.61 & 16.99 & 18.54 & 18.28 & 18.38 & 17.91 \\
\addlinespace[2pt]
\multicolumn{10}{@{}l}{\emph{Exams}} \\
AGIEval & 3 & accuracy & 36.10 & 40.02 & 40.07 & 42.67 & 42.69 & 42.58 & 41.83 \\
AGIEval & 1 & accuracy & 35.28 & 39.92 & 40.07 & 40.63 & 40.54 & 41.14 & 40.20 \\
\addlinespace[2pt]
\multicolumn{10}{@{}l}{\emph{Commonsense}} \\
CommonsenseQA & 5 & accuracy & 66.50 & 66.42 & 66.18 & 67.08 & 67.49 & 66.67 & 67.08 \\
HellaSwag & 5 & accuracy & 70.00 & 70.60 & 70.70 & 70.00 & 70.10 & 70.10 & 70.10 \\
PIQA & 5 & accuracy & 75.50 & 75.20 & 74.90 & 74.30 & 75.20 & 74.90 & 74.80 \\
WinoGrande & 5 & accuracy & 66.61 & 67.96 & 67.72 & 68.27 & 68.35 & 68.90 & 68.75 \\
\addlinespace[2pt]
\multicolumn{10}{@{}l}{\emph{Science QA}} \\
ARC-Challenge & 5 & accuracy & 51.79 & 59.13 & 59.13 & 60.32 & 59.81 & 61.26 & 59.81 \\
ARC-Easy & 5 & accuracy & 78.60 & 80.50 & 79.60 & 81.70 & 82.00 & 80.80 & 80.40 \\
OpenBookQA & 5 & accuracy & 53.40 & 58.40 & 59.00 & 59.80 & 60.60 & 59.00 & 59.60 \\
\addlinespace[2pt]
\multicolumn{10}{@{}l}{\emph{Reading}} \\
CoQA & 0 & F1 & 67.14 & 74.50 & 74.10 & 68.05 & 68.28 & 66.59 & 64.44 \\
SQuAD & 5 & F1 & 82.41 & 87.31 & 88.28 & 87.56 & 87.45 & 88.40 & 88.16 \\
DROP & 5 & F1 & 34.72 & 39.80 & 41.46 & 39.21 & 40.04 & 35.69 & 36.94 \\
\addlinespace[2pt]
\multicolumn{10}{@{}l}{\emph{Open QA}} \\
Jeopardy & 5 & F1 & 63.81 & 65.42 & 65.54 & 64.83 & 64.88 & 64.59 & 65.50 \\
NaturalQuestions & 5 & F1 & 20.84 & 20.83 & 21.48 & 18.77 & 19.02 & 18.77 & 19.28 \\
TriviaQA & 5 & F1 & 55.02 & 56.11 & 55.75 & 55.36 & 55.39 & 54.32 & 54.34 \\
\bottomrule
\end{tabular}
}
\end{table}

\end{document}